\documentclass[runningheads]{llncs}
\usepackage[latin9]{inputenc}
\usepackage{array}
\usepackage{float}
\usepackage{multirow}
\usepackage{amsmath}
\usepackage{amssymb}
\usepackage{graphicx}
\usepackage{xargs}[2008/03/08]

\makeatletter

\newcommand{\noun}[1]{\textsc{#1}}
\providecommand{\tabularnewline}{\\}
\newcommand{\lyxdot}{.}

\usepackage{graphicx}
\usepackage{amsmath,amssymb}
\usepackage{algorithm,algpseudocode}
\usepackage{comment}
\usepackage{color}

\title{Improving Adversarial Robustness by Enforcing Local and Global Compactness}

\titlerunning{Improving Adversarial Robustness by Local and Global Compactness}

\author{Anh Bui\inst{1}\orcidID{0000-0003-4123-2628} \and
Trung Le\inst{1}\orcidID{0000-0003-0414-9067} \and
He Zhao\inst{1}\orcidID{0000-0003-0894-2265} \and 
Paul Montague\inst{2}\orcidID{0000-0001-9461-7471} \and 
Olivier deVel\inst{2}\orcidID{0000-0001-5179-3707} \and 
Tamas Abraham\inst{2}\orcidID{0000-0003-2466-7646} \and 
Dinh Phung\inst{1}\orcidID{0000-0002-9977-8247}}
\authorrunning{A. Bui et al.}
\institute{Monash University, Australia \\
\email{\{tuananh.bui,trunglm,ethan.zhao,dinh.phung\}@monash.edu} \and
Defence Science and Technology Group, Australia \\
\email{\{paul.montague,olivier.devel,tamas.abraham\}@dst.defence.gov.au}}

\makeatother

\begin{document}
\pagestyle{headings} \mainmatter 
\global\long\def\ECCVSubNumber{5686}%

\global\long\def\sidenote#1{\marginpar{\small\emph{{\color{Medium}#1}}}}%

\global\long\def\se{\hat{\text{se}}}%
\global\long\def\interior{\text{int}}%
\global\long\def\boundary{\text{bd}}%
\global\long\def\ML{\textsf{ML}}%
\global\long\def\GML{\mathsf{GML}}%
\global\long\def\HMM{\mathsf{HMM}}%
\global\long\def\support{\text{supp}}%
\global\long\def\new{\text{*}}%
\global\long\def\stir{\text{Stirl}}%
\global\long\def\mA{\mathcal{A}}%
\global\long\def\mB{\mathcal{B}}%
\global\long\def\expect{\mathbb{E}}%
\global\long\def\mF{\mathcal{F}}%
\global\long\def\mK{\mathcal{K}}%
\global\long\def\mH{\mathcal{H}}%
\global\long\def\mX{\mathcal{X}}%
\global\long\def\mZ{\mathcal{Z}}%
\global\long\def\mS{\mathcal{S}}%
\global\long\def\Ical{\mathcal{I}}%
\global\long\def\mT{\mathcal{T}}%
\global\long\def\Pcal{\mathcal{P}}%
\global\long\def\dist{d}%
\global\long\def\HX{\entro\left(X\right)}%
\global\long\def\entropyX{\HX}%
\global\long\def\HY{\entro\left(Y\right)}%
\global\long\def\entropyY{\HY}%
\global\long\def\HXY{\entro\left(X,Y\right)}%
\global\long\def\entropyXY{\HXY}%
\global\long\def\mutualXY{\mutual\left(X;Y\right)}%
\global\long\def\mutinfoXY{\mutualXY}%
\global\long\def\given{\mid}%
\global\long\def\gv{\given}%
\global\long\def\goto{\rightarrow}%
\global\long\def\asgoto{\stackrel{a.s.}{\longrightarrow}}%
\global\long\def\pgoto{\stackrel{p}{\longrightarrow}}%
\global\long\def\dgoto{\stackrel{d}{\longrightarrow}}%
\global\long\def\lik{\mathcal{L}}%
\global\long\def\logll{\mathit{l}}%
\global\long\def\bigcdot{\raisebox{-0.5ex}{\scalebox{1.5}{\ensuremath{\cdot}}}}%
\global\long\def\sig{\textrm{sig}}%
\global\long\def\likelihood{\mathcal{L}}%
\global\long\def\vectorize#1{\mathbf{#1}}%

\global\long\def\vt#1{\mathbf{#1}}%
\global\long\def\gvt#1{\boldsymbol{#1}}%
\global\long\def\idp{\ \bot\negthickspace\negthickspace\bot\ }%
\global\long\def\cdp{\idp}%
\global\long\def\das{}%
\global\long\def\id{\mathbb{I}}%
\global\long\def\idarg#1#2{\id\left\{  #1,#2\right\}  }%
\global\long\def\iid{\stackrel{\text{iid}}{\sim}}%
\global\long\def\bzero{\vt 0}%
\global\long\def\bone{\mathbf{1}}%
\global\long\def\a{\mathrm{a}}%
\global\long\def\ba{\mathbf{a}}%
\global\long\def\b{\mathrm{b}}%
\global\long\def\bb{\mathbf{b}}%
\global\long\def\B{\mathrm{B}}%
\global\long\def\boldm{\boldsymbol{m}}%
\global\long\def\c{\mathrm{c}}%
\global\long\def\C{\mathrm{C}}%
\global\long\def\d{\mathrm{d}}%
\global\long\def\D{\mathrm{D}}%
\global\long\def\N{\mathrm{N}}%
\global\long\def\h{\mathrm{h}}%
\global\long\def\H{\mathrm{H}}%
\global\long\def\bH{\mathbf{H}}%
\global\long\def\K{\mathrm{K}}%
\global\long\def\M{\mathrm{M}}%
\global\long\def\bff{\vt f}%
\global\long\def\bx{\mathbf{\mathbf{x}}}%

\global\long\def\bl{\boldsymbol{l}}%
\global\long\def\s{\mathrm{s}}%
\global\long\def\T{\mathrm{T}}%
\global\long\def\bu{\mathbf{u}}%
\global\long\def\v{\mathrm{v}}%
\global\long\def\bv{\mathbf{v}}%
\global\long\def\bo{\boldsymbol{o}}%
\global\long\def\bh{\mathbf{h}}%
\global\long\def\bs{\boldsymbol{s}}%
\global\long\def\x{\mathrm{x}}%
\global\long\def\bx{\mathbf{x}}%
\global\long\def\bz{\mathbf{z}}%
\global\long\def\hbz{\hat{\bz}}%
\global\long\def\z{\mathrm{z}}%
\global\long\def\y{\mathrm{y}}%
\global\long\def\bxnew{\boldsymbol{y}}%
\global\long\def\bX{\boldsymbol{X}}%
\global\long\def\tbx{\tilde{\bx}}%
\global\long\def\by{\boldsymbol{y}}%
\global\long\def\bY{\boldsymbol{Y}}%
\global\long\def\bZ{\boldsymbol{Z}}%
\global\long\def\bU{\boldsymbol{U}}%
\global\long\def\bn{\boldsymbol{n}}%
\global\long\def\bV{\boldsymbol{V}}%
\global\long\def\bI{\boldsymbol{I}}%
\global\long\def\J{\mathrm{J}}%
\global\long\def\bJ{\mathbf{J}}%
\global\long\def\w{\mathrm{w}}%
\global\long\def\bw{\vt w}%
\global\long\def\bW{\mathbf{W}}%
\global\long\def\balpha{\gvt{\alpha}}%
\global\long\def\bdelta{\boldsymbol{\delta}}%
\global\long\def\bsigma{\gvt{\sigma}}%
\global\long\def\bbeta{\gvt{\beta}}%
\global\long\def\bmu{\gvt{\mu}}%
\global\long\def\btheta{\boldsymbol{\theta}}%
\global\long\def\blambda{\boldsymbol{\lambda}}%
\global\long\def\bgamma{\boldsymbol{\gamma}}%
\global\long\def\bpsi{\boldsymbol{\psi}}%
\global\long\def\bphi{\boldsymbol{\phi}}%
\global\long\def\bpi{\boldsymbol{\pi}}%
\global\long\def\bomega{\boldsymbol{\omega}}%
\global\long\def\bepsilon{\boldsymbol{\epsilon}}%
\global\long\def\btau{\boldsymbol{\tau}}%
\global\long\def\bxi{\boldsymbol{\xi}}%
\global\long\def\realset{\mathbb{R}}%
\global\long\def\realn{\realset^{n}}%
\global\long\def\integerset{\mathbb{Z}}%
\global\long\def\natset{\integerset}%
\global\long\def\integer{\integerset}%

\global\long\def\natn{\natset^{n}}%
\global\long\def\rational{\mathbb{Q}}%
\global\long\def\rationaln{\rational^{n}}%
\global\long\def\complexset{\mathbb{C}}%
\global\long\def\comp{\complexset}%

\global\long\def\compl#1{#1^{\text{c}}}%
\global\long\def\and{\cap}%
\global\long\def\compn{\comp^{n}}%
\global\long\def\comb#1#2{\left({#1\atop #2}\right) }%
\global\long\def\nchoosek#1#2{\left({#1\atop #2}\right)}%
\global\long\def\param{\vt w}%
\global\long\def\Param{\Theta}%
\global\long\def\meanparam{\gvt{\mu}}%
\global\long\def\Meanparam{\mathcal{M}}%
\global\long\def\meanmap{\mathbf{m}}%
\global\long\def\logpart{A}%
\global\long\def\simplex{\Delta}%
\global\long\def\simplexn{\simplex^{n}}%
\global\long\def\dirproc{\text{DP}}%
\global\long\def\ggproc{\text{GG}}%
\global\long\def\DP{\text{DP}}%
\global\long\def\ndp{\text{nDP}}%
\global\long\def\hdp{\text{HDP}}%
\global\long\def\gempdf{\text{GEM}}%
\global\long\def\rfs{\text{RFS}}%
\global\long\def\bernrfs{\text{BernoulliRFS}}%
\global\long\def\poissrfs{\text{PoissonRFS}}%
\global\long\def\grad{\gradient}%
\global\long\def\gradient{\nabla}%
\global\long\def\partdev#1#2{\partialdev{#1}{#2}}%
\global\long\def\partialdev#1#2{\frac{\partial#1}{\partial#2}}%
\global\long\def\partddev#1#2{\partialdevdev{#1}{#2}}%
\global\long\def\partialdevdev#1#2{\frac{\partial^{2}#1}{\partial#2\partial#2^{\top}}}%
\global\long\def\closure{\text{cl}}%
\global\long\def\cpr#1#2{\Pr\left(#1\ |\ #2\right)}%
\global\long\def\var{\text{Var}}%
\global\long\def\Var#1{\text{Var}\left[#1\right]}%
\global\long\def\cov{\text{Cov}}%
\global\long\def\Cov#1{\cov\left[ #1 \right]}%
\global\long\def\COV#1#2{\underset{#2}{\cov}\left[ #1 \right]}%
\global\long\def\corr{\text{Corr}}%
\global\long\def\sst{\text{T}}%
\global\long\def\SST{\sst}%
\global\long\def\ess{\mathbb{E}}%

\global\long\def\Ess#1{\ess\left[#1\right]}%
\newcommandx\ESS[2][usedefault, addprefix=\global, 1=]{\underset{#2}{\ess}\left[#1\right]}%
\global\long\def\fisher{\mathcal{F}}%

\global\long\def\bfield{\mathcal{B}}%
\global\long\def\borel{\mathcal{B}}%
\global\long\def\bernpdf{\text{Bernoulli}}%
\global\long\def\betapdf{\text{Beta}}%
\global\long\def\dirpdf{\text{Dir}}%
\global\long\def\gammapdf{\text{Gamma}}%
\global\long\def\gaussden#1#2{\text{Normal}\left(#1, #2 \right) }%
\global\long\def\gauss{\mathbf{N}}%
\global\long\def\gausspdf#1#2#3{\text{Normal}\left( #1 \lcabra{#2, #3}\right) }%
\global\long\def\multpdf{\text{Mult}}%
\global\long\def\poiss{\text{Pois}}%
\global\long\def\poissonpdf{\text{Poisson}}%
\global\long\def\pgpdf{\text{PG}}%
\global\long\def\wshpdf{\text{Wish}}%
\global\long\def\iwshpdf{\text{InvWish}}%
\global\long\def\nwpdf{\text{NW}}%
\global\long\def\niwpdf{\text{NIW}}%
\global\long\def\studentpdf{\text{Student}}%
\global\long\def\unipdf{\text{Uni}}%
\global\long\def\transp#1{\transpose{#1}}%
\global\long\def\transpose#1{#1^{\mathsf{T}}}%
\global\long\def\mgt{\succ}%
\global\long\def\mge{\succeq}%
\global\long\def\idenmat{\mathbf{I}}%
\global\long\def\trace{\mathrm{tr}}%
\global\long\def\argmax#1{\underset{_{#1}}{\text{argmax}} }%
\global\long\def\argmin#1{\underset{_{#1}}{\text{argmin}\ } }%
\global\long\def\diag{\text{diag}}%
\global\long\def\norm{}%
\global\long\def\spn{\text{span}}%
\global\long\def\vtspace{\mathcal{V}}%
\global\long\def\field{\mathcal{F}}%
\global\long\def\ffield{\mathcal{F}}%
\global\long\def\inner#1#2{\left\langle #1,#2\right\rangle }%
\global\long\def\iprod#1#2{\inner{#1}{#2}}%
\global\long\def\dprod#1#2{#1 \cdot#2}%
\global\long\def\norm#1{\left\Vert #1\right\Vert }%
\global\long\def\entro{\mathbb{H}}%
\global\long\def\entropy{\mathbb{H}}%
\global\long\def\Entro#1{\entro\left[#1\right]}%
\global\long\def\Entropy#1{\Entro{#1}}%
\global\long\def\mutinfo{\mathbb{I}}%
\global\long\def\relH{\mathit{D}}%
\global\long\def\reldiv#1#2{\relH\left(#1||#2\right)}%
\global\long\def\KL{KL}%
\global\long\def\KLdiv#1#2{\KL\left(#1\parallel#2\right)}%
\global\long\def\KLdivergence#1#2{\KL\left(#1\ \parallel\ #2\right)}%
\global\long\def\crossH{\mathcal{C}}%
\global\long\def\crossentropy{\mathcal{C}}%
\global\long\def\crossHxy#1#2{\crossentropy\left(#1\parallel#2\right)}%
\global\long\def\breg{\text{BD}}%
\global\long\def\lcabra#1{\left|#1\right.}%
\global\long\def\lbra#1{\lcabra{#1}}%
\global\long\def\rcabra#1{\left.#1\right|}%
\global\long\def\rbra#1{\rcabra{#1}}%

\global\long\def\model{\text{ADR}}%
\maketitle
\begin{abstract}
The fact that deep neural networks are susceptible to crafted perturbations
severely impacts the use of deep learning in certain domains of application.
Among many developed defense models against such attacks, adversarial
training emerges as the most successful method that consistently resists
a wide range of attacks. In this work, based on an observation from
a previous study that the representations of a clean data example
and its adversarial examples become more divergent in higher layers
of a deep neural net, we propose the Adversary Divergence Reduction
Network which enforces local/global compactness and the clustering
assumption over an intermediate layer of a deep neural network. We
conduct comprehensive experiments to understand the isolating behavior
of each component (i.e., local/global compactness and the clustering
assumption) and compare our proposed model with state-of-the-art adversarial
training methods. The experimental results demonstrate that augmenting
adversarial training with our proposed components can further improve
the robustness of the network, leading to higher unperturbed and adversarial
predictive performances. 

\keywords{Adversarial Robustness, Local Compactness, Global Compactness, Clustering assumption}
\end{abstract}

\section{Introduction}

Despite the great success of deep neural nets, they are reported to
be susceptible to crafted perturbations \cite{szegedy2013intriguing,goodfellow2014explaining},
even state-of-the-art ones. Accordingly, many defense models have
been developed, notably \cite{madry2017towards,Zhang2019theoretically,xie2019feature,qin2019adversarial}.
Recently, the work of \cite{athalye2018obfuscated} undertakes an
in-depth study of neural network defense models and conduct comprehensive
experiments on a complete suite of defense techniques, which has lead
to postulating one common reason why many defenses provide apparent
robustness against  gradient-based attacks, namely \emph{obfuscated gradients}. 

According to the above study, adversarial training with Projected
Gradient Descent (PGD) \cite{madry2017towards} is one of the most
successful and widely-used defense techniques that remained consistently
resilient against attacks, which has inspired many recent advances
including Adversarial Logit Pairing (ALP) \cite{kannan2018adversarial},
Feature Denoising \cite{xie2019feature}, Defensive Quantization \cite{lin2019defensive},
Jacobian Regularization \cite{jakubovitz2018improving}, Stochastic
Activation Pruning \cite{dhillon2018stochastic}, and Adversarial
Training Free \cite{shafahi2019adversarial}.

In this paper, we propose to build robust classifiers against adversarial
examples by learning better representations in the intermediate space.
Given an image classifier based on a multi-layer neural net, conceptually,
we divide the network into two parts with an intermediate layer: the
generator network from the input layer to the intermediate layer and
the classifier network from the intermediate layer to the output prediction
layer. The output of the generator network (i.e., the intermediate
layer) is the intermediate representation of the input image, which
is fed to the classifier network to make prediction. For image classifiers,
an adversarial example is usually generated by adding small perturbations
to a clean image. The adversarial example may look very similar to
the original image but leads to significant changes to the prediction
of the classifier. It has been observed that in deep neural networks,
the representations of a clean data example and its adversarial example
might become very diverge in the intermediate space, although\emph{
}their representations are proximal in the data space \cite{xie2019feature}.
Due to the above divergence in the intermediate space, a classifier
may be hard to predict the same class of the adversarial and real
images. Inspired by this observation, we propose to learn better representations
that reduce the above divergence in the intermediate space, so as
to enhance the classifier robustness against adversarial examples.

In particular, we propose an enhanced adversarial training framework
that imposes the $\emph{local and global compactness}$ properties
on the intermediate representations, to build more robust classifiers
against adversarial examples. Specifically, by explicitly strengthening
local compactness, we enforce the intermediate representations output
from the generator of a clean image and its adversarial examples to
be as proximal as possible. In this way, the classifier network is
less easy to be misled by the adversarial examples. However, enforcing
the local compactness itself may not be sufficient to guarantee a
robust defense model as the representations might be encouraged to
globally spread out in the intermediate space, significantly hurting
accuracies on both clean and adversarial images. To address this,
we further propose to impose global compactness to encourage the representations
of examples in the same class to be proximal yet those in different
classes to be more distant. Finally, to increase the generalization
capacity of the deep network and reduce the misclassification of adversarial
examples, our framework enjoys the flexibility to incorporate the
clustering assumption \cite{chapelle2005semi}, which aims to force
the decision boundary of a classifier to lie in the gap between clusters
of different classes. By collaboratively incorporating the above three
properties, we are able to learn better intermediate representations,
which help to boost the adversarial robustness of classifiers. Intuitively,
we name our proposed framework to the \emph{Adversary Divergence Reduction
Network} (ADR).

To comprehensively exam the proposed framework, we conduct extensive
experiments to investigate the influence of each component (i.e.,
local/global compactness and the clustering assumption), visualize
the smoothness of the loss surface of our robust model, and compare
our proposed ADR method with several state-of-the-art adversarial
defenses. The experimental results consistently show that our proposed
method can further improve over others in terms of better adversarial
and clean predictive performances. The contributions of this work
are summarized as follows: 
\begin{itemize}
\item We propose the local and global compactness properties on the intermediate
space to enforce the better representations, which lead to more robust
classifiers;
\item We incorporate our local and global compactness with clustering assumption
to further enhance adversarial robustness; 
\item We plug the above three components into an adversarial training framework
to introduce our Adversary Divergence Reduction Network; 
\item We extensively analyze the proposed framework and compare it with
state-of-the-art adversarial training methods to verify its effectiveness. 
\end{itemize}

\section{Related works}

\subsubsection{Adversarial training defense}

Adversarial training can be traced back to \cite{goodfellow2014explaining},
in which models were challenged by producing adversarial examples
and incorporating them into training data. The adversarial examples
could be the worst-case examples (i.e., $x_{a}\triangleq\text{argmax}{}_{x'\in B_{\varepsilon}\left(x\right)}\ell\left(x',y,\theta\right)$)
\cite{goodfellow2014explaining} or most divergent examples (i.e.,
$x_{a}\triangleq\text{argmax}{}_{x'\in B_{\varepsilon}\left(x\right)}D_{KL}\left(h_{\theta}\left(x'\right)\mid\mid h_{\theta}\left(x\right)\right)$)
\cite{Zhang2019theoretically} where $D_{KL}$ is the Kullback-Leibler
divergence and $h_{\theta}$ is the current model. The quality of
the adversarial training defense crucially depends on the strength
of the injected adversarial examples -- e.g., training on non-iterative
adversarial examples obtained from FGSM or Rand FGSM (a variant of
FGSM where the initial point is randomised) are not robust to iterative
attacks, for example PGD \noun{\cite{madry2017towards} }or BIM \cite{kurakin2016adversarial}.

Although many defense models were broken by \cite{athalye2018obfuscated},
the adversarial training with PGD \cite{madry2017towards} was among
the few that were resilient against attacks. Many defense models were
developed based on adversarial examples from a PGD attack or attempts
made to improve and scale up the PGD adversarial training. Notable
examples include Adversarial Logit Pairing (ALP) \cite{kannan2018adversarial},
Feature Denoising \cite{xie2019feature}, Defensive Quantization \cite{lin2019defensive},
Jacobian Regularization \cite{jakubovitz2018improving}, Stochastic
Activation Pruning \cite{dhillon2018stochastic}, and Adversarial
Training for Free \cite{shafahi2019adversarial}.

\subsubsection{Defense with a latent space }

These works utilized a latent space to enable adversarial defense,
notably \cite{jalal2017robust}. DefenseGAN \cite{samangouei2018defense}
and PixelDefense \cite{song2017pixeldefend} use a generator (i.e.,
a pretrained WS-GAN \cite{gulrajani2017improved} for DefenseGAN and
a PixelCNN \cite{oord2016pixel} for PixelDefense) together with the
latent space to find a denoised version of an adversarial example
on the data manifold. These works were criticized by \cite{athalye2018obfuscated}
as being easy to attack and impossible to work within the case of
the CIFAR-10 dataset. Jalal et al. \cite{jalal2017robust} proposed
an overpowered attack method to efficiently attack both DefenseGAN
and PixelDefense and subsequently injected those adversarial examples
to train the model. Though that work was proven to work well with
simple datasets including MNIST and CelebA, no experiments were conducted
on more complex datasets including, for example, CIFAR-10.

\section{Proposed method}

In what follows, we present our proposed method, named the \emph{Adversary
Divergence Reduction Network} (ADR). As shown in the previous study
\cite{xie2019feature}, although an adversarial example $x_{a}$ and
its corresponding clean example $x$ are in close proximity in the
data space (i.e., differ by a small perturbation), when brought forward
up to the higher layers in a deep neural network, their representations
become markedly more divergent, hence causing different prediction
results. Inspired by this observation, we propose imposing local compactness
for those representations in an intermediate layer of a neural network.
The key idea is to enforce that the representations of an adversarial
example and its clean counterpart be as proximal as possible, hence
reducing the chance of misclassifying them. Moreover, we observe that
enforcing the local compactness itself is not sufficient to guarantee
a robust defense model as this enforcement might encourage representations
to globally spread out across the intermediate space (i.e., the space
induced by the intermediate representations), significantly hurting
both adversarial and clean performances. To address this, we propose
to impose global compactness for the intermediate representations
such that representations of examples that belong in the same class
are proximal and those in different classes are more distant. Finally,
to increase the generalization capacity of the deep network and reduce
the misclassification of adversarial examples, we propose to apply
the clustering assumption \cite{chapelle2005semi} which aims to force
the decision boundary to lie in the gap between clusters of different
classes, hence increasing the chance for adversarial examples to be
correctly classified.

\subsection{Local compactness }

Local compactness, which aims to reduce the divergence between the
representations of an adversarial example and its clean example in
an intermediate layer, is one of the key aspects of our proposed method.
Let us denote our deep neural network by $h_{\theta}\left(\cdot\right)$,
which decomposes into $h_{\theta}\left(\cdot\right)=g_{\theta}\left(f_{\theta}\left(\cdot\right)\right)$
where the first (generator) network $f_{\theta}$ maps the data examples
onto an intermediate layer where we enforce the compactness constraints.
The following (classifier) network $g_{\theta}$ maps the intermediate
representations to the prediction output. For local compactness, given
a clean data example $x$, denote $\mathcal{A}_{\varepsilon}$ as
a stochastic adversary that renders adversarial examples for $x$
as $x_{a}\sim\mathcal{A}_{\varepsilon}\left(x\right)$ in a ball $B_{\varepsilon}\left(x\right)=\left\{ x':\norm{x-x'}<\varepsilon\right\} $,
our aim is to compress the representations of $x$ and $x_{a}$ in
the intermediate layer by minimizing 
\begin{equation}
\mathcal{{L}}_{\text{com}}^{\text{lc}}=\mathbb{{E}}_{x\sim\mathcal{D}_{x}}\left[\mathbb{{E}}_{x_{a}\sim\mathcal{A}_{\varepsilon}\left(x\right)}\left[\norm{f_{\theta}(x)-f_{\theta}(x_{a})}_{p}\right]\right]\label{eq:local}
\end{equation}
where we use $\mathcal{D}_{x}=\left\{ x_{1},...,x_{N}\right\} $ to
represent both training examples and the corresponding empirical distribution
and $\norm{\cdot}_{p}$ with $p=1,2,\infty$ to specify the $p$-norm.

\subsection{Global compactness}

For global compactness, we want the representations of data examples
in the same class to be closer and data examples in different classes
to be more separate. As demonstrated later, global compactness in
conjunction with the clustering assumption helps increase the margin
of a data example (i.e., the distance from that data example to the
decision boundary), hence boosting the generalization capacity of
the classifier network and adversarial robustness.

More specifically, given two examples $\left(x_{i},y_{i}\right)$
and $\left(x_{j},y_{j}\right)$ drawn from the empirical distribution
over $\mathcal{D}_{x,y}=\left\{ \left(x_{1},y_{1}\right),...,\left(x_{N},y_{N}\right)\right\} $
where the label $y_{k}\in\left\{ 1,2,...,M\right\} $, we compute
the weight $w_{ij}$ for this pair as follows: 
\begin{equation}
w_{ij}=\frac{\alpha-\mathbb{I}_{y_{i}\neq y_{j}}}{\alpha}=\begin{cases}
1 & \text{if}\,y_{i}=y_{j}\\
\frac{\alpha-1}{\alpha} & \text{otherwise}
\end{cases}
\end{equation}
where $\mathbb{I}_{S}$ is the indicator function which returns 1
if $S$ holds and $0$ otherwise. We consider $\alpha\in(0,1)$, yielding
$w_{ij}<0$ if $y_{i}\neq y_{j}$ and $w_{ij}>0$ if otherwise.

We enforce global compactness by minimizing 
\begin{equation}
\mathcal{{L}}_{\text{com}}^{\text{gb}}=\mathbb{{E}}_{\left(x_{i},y_{i}\right),\left(x_{j},y_{j}\right)\sim\mathcal{D}_{x,y}}\left[w_{ij}\norm{f_{\theta}(x_{i})-f_{\theta}(x_{j})}_{p}\right]\label{eq:global}
\end{equation}
where we overload the notation $\mathcal{D}_{x,y}$ to represent the
empirical distribution over the training set, which implies that the
intermediate representations $f_{\theta}(x_{i})\text{ }$ and $f_{\theta}(x_{j})$
are encouraged to be closer if $y_{i}=y_{j}$ and to be separate if
$y_{i}\neq y_{j}$ for a global compact representation.

Note that in our experiment, we set $\alpha=0.99$, yielding $w_{ij}\in\{1,-0.01\}$,
and calculate global compactness with each random minibatch. 

\subsection{Clustering assumption and label supervision}

At this stage, we have achieved compact intermediate representations
for the clean data and adversarial examples obtained from a stochastic
adversary $\mathcal{A}_{\varepsilon}$. Our next step is to enforce
some constraints on the subsequent classifier network $g_{\theta}$
to further exploit this compact representation for improving adversarial
robustness. The first constraint we impose on the classifier network
$g_{\theta}$ is that this should classify both clean data and adversarial
examples correctly by minimizing
\begin{equation}
\mathcal{L}_{\text{c}}=\mathbb{E}_{\left(x,y\right)\sim D_{x,y}}\Biggl[\mathbb{E}_{x_{a}\sim\mathcal{A}_{\varepsilon}\left(x\right)}\left[\ell\left(h_{\theta}\left(x_{a}\right),y\right)\right]+\ell\left(h_{\theta}\left(x\right),y\right)\Biggr]\label{eq:label supervision}
\end{equation}
where $\ell$ is the cross-entropy loss function.

In addition to this label supervision, the second constraint we impose
on the classifier network $g_{\theta}$ is the clustering assumption
\cite{chapelle2005semi}, which states that the decision boundary
of $g_{\theta}$ in the intermediate space should not break into any
high density region (or cluster) of data representations in the intermediate
space, forcing the boundary to lie in gaps formed by those clusters.
The clustering assumption when combined with the global compact representation
property should increase the data example margin (i.e., the distance
from that data example to the decision boundary). If this is further
combined with the fact that the representations of adversarial examples
are compressed into the representation of its clean data example ({\em
i.e.} local compactness) this should also reduce the chance that
adversarial examples are misclassified. To enforce the clustering
assumption, inspired by \cite{shu2018dirt}, we encourage the classifier
confidence by minimizing the conditional entropy and maintain classifier
smoothness using Virtual Adversarial Training (VAT) \cite{VAT}, respectively: 

\begin{equation}
\mathcal{{L}}_{\text{conf}}=\mathbb{{E}}_{x\sim D_{x}}\Biggl[\mathbb{{E}}_{x_{a}\sim\mathcal{A}_{\varepsilon}\left(x\right)}\left[-h_{\theta}(x_{a})^{T}\log h_{\theta}(x_{a})\right]-h_{\theta}(x)^{T}\log h_{\theta}(x)\Biggr]\label{eq:conf}
\end{equation}

\begin{equation}
\mathcal{{L}}_{\text{smt}}=\mathbb{{E}}_{x\sim D_{x}}\left[\mathbb{{E}}_{x_{a}\sim\mathcal{A}_{\varepsilon}\left(x\right)}\left[D_{KL}\left(h_{\theta}(x)\Vert h_{\theta}(x_{a})\right)\right]\right]\label{eq:smt}
\end{equation}

\subsection{Generating adversarial examples\label{subsec:adv_framework} }

We can use any adversarial attack algorithm to define the adversary
$\mathcal{A}_{\varepsilon}$. For example, Madry et al. \cite{madry2017towards}
proposed to find the worst-case examples $x_{a}\triangleq\text{argmax}{}_{x'\in B_{\varepsilon}\left(x\right)}\ell\left(x',y,\theta\right)$
using PGD, while Zhang et al. \cite{Zhang2019theoretically} aimed
to find the most divergent examples $x_{a}\triangleq\text{argmax}{}_{x'\in B_{\varepsilon}\left(x\right)}D_{KL}\left(h_{\theta}\left(x'\right)\mid\mid h_{\theta}\left(x\right)\right).$
By enforcing local/global compactness over the adversarial examples
obtained by $\mathcal{A}_{\varepsilon}$, we make them easier to be
trained with the label supervision loss in Eq. (\ref{eq:label supervision}),
hence eventually improving adversarial robustness. The quality of
adversarial examples obviously affects to the overall performance,
however, in the experimental section, we empirically prove that our
proposed components can boost the robustness of the adversarial training
frameworks of interest.

\begin{figure}
\begin{centering}
\includegraphics[width=0.5\columnwidth]{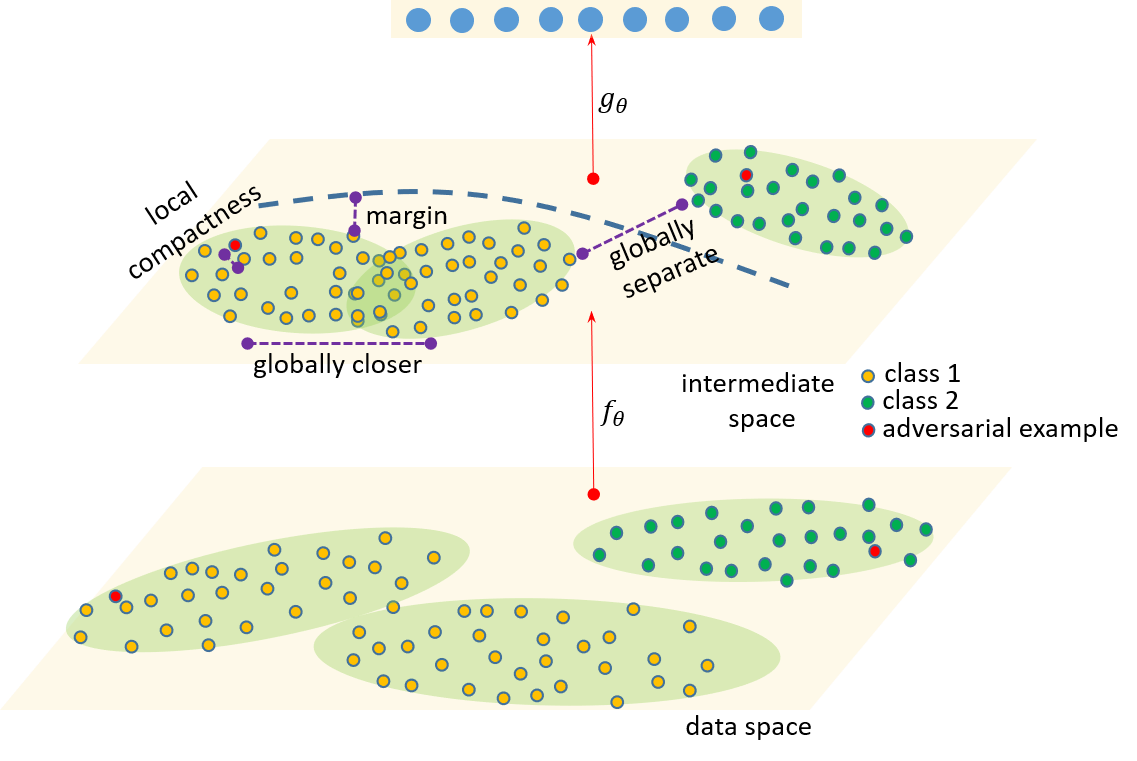}
\par\end{centering}
\caption{Overview of Adversary Divergence Reduction Network. The local/global
compactness and clustering assumption are intended to improve adversarial
robustness.\label{fig:overview}}
\end{figure}

\subsection{Putting it all together \label{subsec:Putting-Altogether}}

We combine the relevant terms regarding local/global compactness,
label supervision, and the clustering assumption and arrive at the
following optimization problem: 
\begin{equation}
\min_{\theta}\,\mathcal{L}\triangleq\mathcal{L}_{\text{c}}+\lambda_{\text{com}}^{\text{lc}}\mathcal{L}_{\text{com}}^{\text{lc}}+\lambda_{\text{com}}^{\text{gb}}\mathcal{L}_{\text{com}}^{\text{gb}}+\lambda_{\text{conf}}\mathcal{L}_{\text{conf}}+\lambda_{\text{smt}}\mathcal{L}_{\text{smt}}\label{eq:total}
\end{equation}
where $\lambda_{\text{com}}^{\text{lc}},\lambda_{\text{com}}^{\text{gb}},\lambda_{\text{conf}},\text{ and }\lambda_{\text{smt}}$
are non-negative trade-off parameters.

In Figure \ref{fig:overview}, we illustrate how the three components,
namely local/global compactness, label supervision, and the clustering
assumption can mutually collaborate to improve adversarial robustness.
The representations of data examples via the network $f_{\theta}$
are enforced to be locally/globally compact, whereas the position
of the decision boundary of the classifier network $g_{\theta}$ in
the intermediate space is enforced using the clustering assumption.
Ideally, with the clustering assumption, the decision boundary of
$g_{\theta}$ preserves the cluster structure in the intermediate
space and when combined with label supervision training ensures clusters
in a class remain completely inside the decision region for this class.
Moreover, global compactness encourages clusters of a class to be
closer and those of different classes to be more separate. As a result,
the decision boundary of $g_{\theta}$ lies in the gaps among clusters
as well as with a sufficiently large margin for the data examples.
Finally, local compactness requires adversarial examples to stay closer
to their corresponding clean data example, hence reducing the chance
of misclassifying them and therefore improving adversarial robustness.

\paragraph{\textbf{Comparison with the contrastive learning. }}

Interestingly, the contrastive learning \cite{chen2020simple,he2020momentum}
and our proposed method aim to learn better representations by the
principle of enforcing similar elements to be equal and dissimilar
elements to be different. However, the contrastive learning works
on an instance level, which enforces the representation of an image
to be proximal with those of its transformations and to be distant
with those of any other images. On the other hand, our method works
on a class level, which enforces the intermediate representations
of each class to be compact and well separated with those in other
classes. Therefore, our method and the contrastive learning complement
each other and intuitively improve both visual representation and
adversarial robustness when combining together.

\section{Experiments}

In this section, we first introduce the general setting for our experiments
regarding datasets, model architecture, optimization scheduler, and
adversary attackers. Second, we compare our method with adversarial
training with PGD, namely ADV \cite{madry2017towards} and TRADES
\cite{Zhang2019theoretically}. We employ either ADV or TRADES as
the stochastic adversary $\mathcal{A}$ for our $\model$ and demonstrate
that, when enhanced with local/global compactness and the clustering
assumption, we can improve these state-of-the-art adversarial training
methods.

Specifically, we begin this section with an ablation study to investigate
the model behaviors and the influence of each component, namely local
compactness, global compactness, and the clustering assumption, on
adversarial performance. In addition, we visualize the smoothness
of the loss surface of our model to understand why it can defend well.
Finally, we undertake experiments on the MNIST and CIFAR-10 datasets
to compare our $\model$ with both ADV and TRADES.

\subsection{Experimental setting }

\paragraph{\textbf{General setting}}

We undertook experiments on both the MNIST \cite{lecun1998gradient}
and CIFAR-10 \cite{krizhevsky2009learning} datasets. The inputs were
normalized to $[0,1]$. For the CIFAR-10 dataset, we apply random
crops and random flips as describe in \cite{madry2017towards} during
training. For the MNIST dataset, we used the standard CNN architecture
with three convolution layers and three fully connected layers described
in \cite{carlini2017towards}. For the CIFAR-10 dataset, we used two
architectures in which one is the standard CNN architecture described
in \cite{carlini2017towards} and another is the ResNet architecture
used in \cite{madry2017towards}. We note that there is a serious
overfitting problem on the CIFAR-10 dataset as mentioned in \cite{carlini2017towards}.
In our setting, with the standard CNN architecture, we eventually
obtained a $98\%$ training accuracy, but only a $75\%$ testing accuracy.
With the ResNet architecture, we used the strategy from \cite{madry2017towards}
to adjust the learning rate when training to reduce the gap between
the training and validation accuracies. For the MNIST dataset, a drop-rate
equal to 0.1 at epochs 55, 75, and 90 without weight decay was employed.
For the CIFAR-10 dataset, the drop-rate was set to 0.1 at epochs 100
and 150 with weight decay equal to $2\times10^{-4}$. We use a momentum-based
SGD optimizer for the training of the standard CNN for the MNIST dataset
and the ResNet for the CIFAR-10 dataset, while using the Adam optimizer
for training the standard CNN on the CIFAR-10 one.  The hyperparameters
setting can be found in the supplementary material. 

\paragraph{\textbf{Choosing the intermediate layer. }}

The intermediate layer for enforcing compactness constraints immediately
follows on from the last convolution layer for the standard CNN architecture
and from the penultimate layer for the ResNet architecture.  Moreover,
we provide an additional ablation study to investigate the importance
of choosing the intermediate layer which can be found in the supplementary
material.

\paragraph{\textbf{Attack methods}}

We use PGD to challenge the defense methods in this paper. Specifically,
the setting for the MNIST dataset is PGD-40 (i.e., PGD with 40 steps)
with the distortion bound $\varepsilon$ increasing from $0.1$ to
$0.7$ and step size $\eta\in\{0.01,0.02\}$, while that for CIFAR-10
is PGD-20 with $\varepsilon$ increasing from $0.0039\,(\approx1/255)$
to $0.11\,(\approx28/255)$ and step size $\eta\in\{0.0039,0.007\}$.
The distortion metric is $l_{\infty}$ for all attacks. For the adversarial
training, we use $k=10$ for CIFAR10 and $k=20$ for MNIST  for all
defense methods.

\paragraph{\textbf{}}

\paragraph{\textbf{Non-targeted and multi-targeted attack scenarios}}

We used two types of attack scenarios, namely non-targeted and multi-targeted
attacks. The non-targeted attack derives adversarial examples by maximizing
the loss w.r.t. its clean data label, whilst the multi-targeted attack
is undertaken by performing simultaneously  targeted attack for all
possible data labels. The multi-targeted attack is considered to be
successful if any individual targeted attack on each target label
is successful. While the non-targeted attack considers only one direction
of the gradient, the multi-targeted attack takes multi-directions
of gradient into account, which guarantees to get better local optimum.

\subsection{Experimental results }

In this section, we first conduct an ablation study using the MNIST
dataset in order to investigate how the different components (local
compactness, global compactness, and the clustering assumption) contribute
to adversarial robustness. We then conduct experiments on the MNIST
and CIFAR-10 datasets to compare our proposed method with ADV and
TRADES. Further evaluation can be found in the supplementary material.

\subsubsection{Ablation study}

We first study how each proposed component contributes to adversarial
robustness. We use adversarial training with PGD as the baseline model
and experiment on the MNIST dataset. Recall that our method consists
of three components: the local compactness loss $\mathcal{L}_{\text{com}}^{\text{lc}}$,
the global compactness loss $\mathcal{{L}}_{\text{com}}^{\text{gb}}$,
and the clustering assumption loss which combines $\{\mathcal{{L}}_{\text{smt}}+\mathcal{{L}}_{\text{conf}}\}$
. In this experiment, we simply set the trade-off parameters $\lambda_{\text{com}}^{\text{lc}},\lambda_{\text{com}}^{\text{gb}},\lambda_{\text{smt}}=\lambda_{\text{conf}}=\lambda_{\text{ca}}$
to $0/1$ to deactivate/activate the corresponding component. We consider
two metrics: the natural accuracy (i.e., the clean accuracy) and the
robustness accuracy to evaluate a defense method. The natural accuracy
is that evaluated on the clean test images, while the robustness accuracy
is that evaluated on adversarial examples generated by attacking the
clean test images. It is noteworthy that for many existing defense
methods, improving robustness accuracy usually harms natural accuracy.
Therefore, our proposed method aims to reach a better trade-off between
the two metrics.

\begin{figure*}[htpb]

\begin{minipage}[t]{0.5\columnwidth}%
\begin{figure}[H]
\begin{centering}
\includegraphics[width=1\columnwidth]{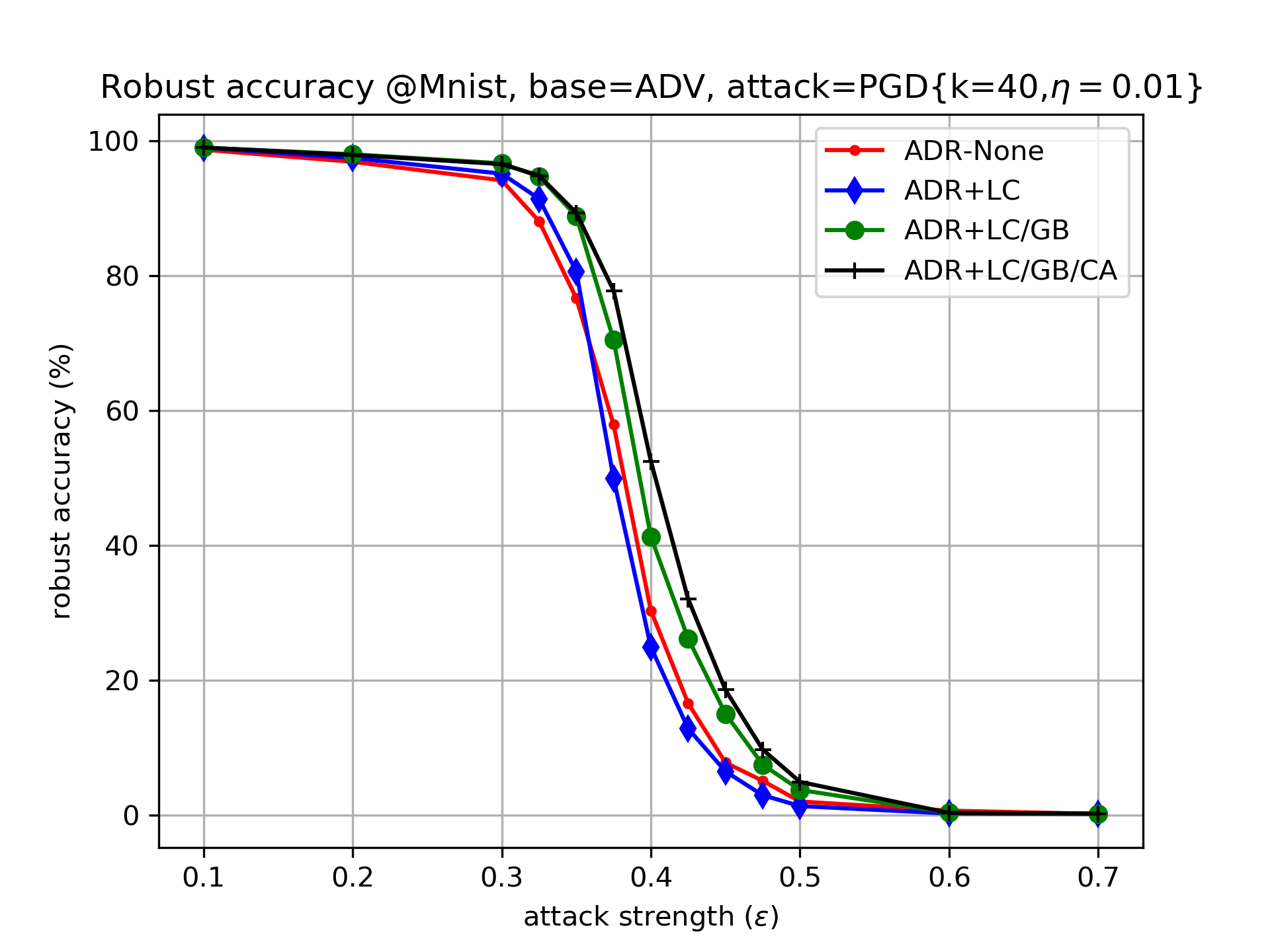}
\par\end{centering}
\caption{Variation of the robustness accuracies under different attack strengths.
The base model is ADV (ADR-None).\label{fig:ablation-mnist-adv}}

\end{figure}
\end{minipage}\hfill{}%
\begin{minipage}[t]{0.45\columnwidth}%
\begin{table}[H]
\begin{centering}
\caption{Results of the PGD-40 attack on the MNIST dataset for the base ADV
model together with its variants with the different components (LC
= local compactness, GB = global compactness, CA = clustering assumption)
and $\varepsilon=0.325$.\label{tab:ablation-mnist-adv}}
\par\end{centering}
\begin{centering}
\begin{tabular}{ccc}
\hline 
 & Nat. acc. & Rob. acc.\tabularnewline
\hline 
Standard model & 99.5\% & 0.84\%\tabularnewline
ADR-None (ADV)\footnote{The performance of ADV is lower than that in \cite{madry2017towards}
because of the difference of the attack strength and model architecture} & 99.27\% & 88.1\%\tabularnewline
ADR+LC & 99.41\% & 91.43\%\tabularnewline
ADR+LC/GB & 99.35\% & 94.52\%\tabularnewline
ADR+LC/GB/CA & 99.36\% & 94.96\%\tabularnewline
\hline 
\end{tabular}
\par\end{centering}
\end{table}
\end{minipage}

\end{figure*}

Table \ref{tab:ablation-mnist-adv} shows the results for the PGD
attack with $k=40,\,\varepsilon=0.325,\text{ and }\eta=0.01$. We
note that ADR-None is our base model without any additional components.
The base model can be any adversarial training based method, e.g.,
ADV or TRADES. Without loss of generality, we use ADV as the base
model, i.e., ADR-None. By gradually combining the proposed additional
components with ADR-None we produce several variants of ADR (e.g.,
ADR+LC is ADR-None together with the local compactness component).
Since the standard model was trained without any defense mechanism,
its natural accuracy is high at $99.5\%$ whereas the robustness accuracy
is very poor at $0.88\%$, indicating its vulnerability to adversarial
attacks. Regarding the variants of our proposed models, those with
additional components generally achieve higher robustness accuracies
compared with ADR-None (i.e. ADV), without hurting the natural accuracy.
In addition, the robust accuracy was significantly improved with global
compactness and the clustering assumption terms.

We also evaluate the metrics of interest with different attack strength
by increasing the distortion boundary $\varepsilon$ as shown in Figure
\ref{fig:ablation-mnist-adv}. By just adding a single local compactness
component, our method can improve the base model (ADV or ADR-None)
for attacks with strength $\varepsilon\leq0.35$. By adding the global
compactness component, our method can significantly improve over the
base model, especially for stronger attacks. Recall that as we generate
adversarial examples from the PGD attack with $k=20,\varepsilon_{d}=0.3,\eta=0.01$
to train the defense models, is is unsurprised to see a model defends
well with $\varepsilon\leq0.3$. Interestingly, by adding our components,
our defense methods can also achieve reasonably good robustness accuracy
of $80\%$, even when $\varepsilon$ varies from 0.34 to $0.37$,
indicating the better generality of our methods. 

\begin{figure*}
\begin{centering}
\includegraphics[width=0.35\textwidth]{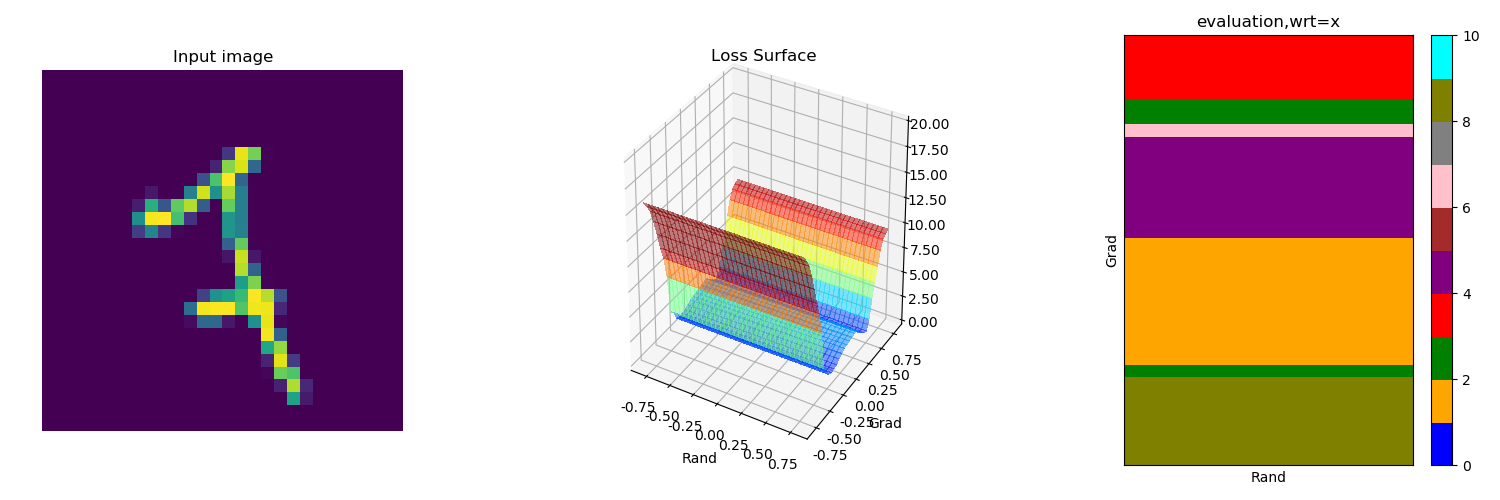}~~\includegraphics[width=0.35\textwidth]{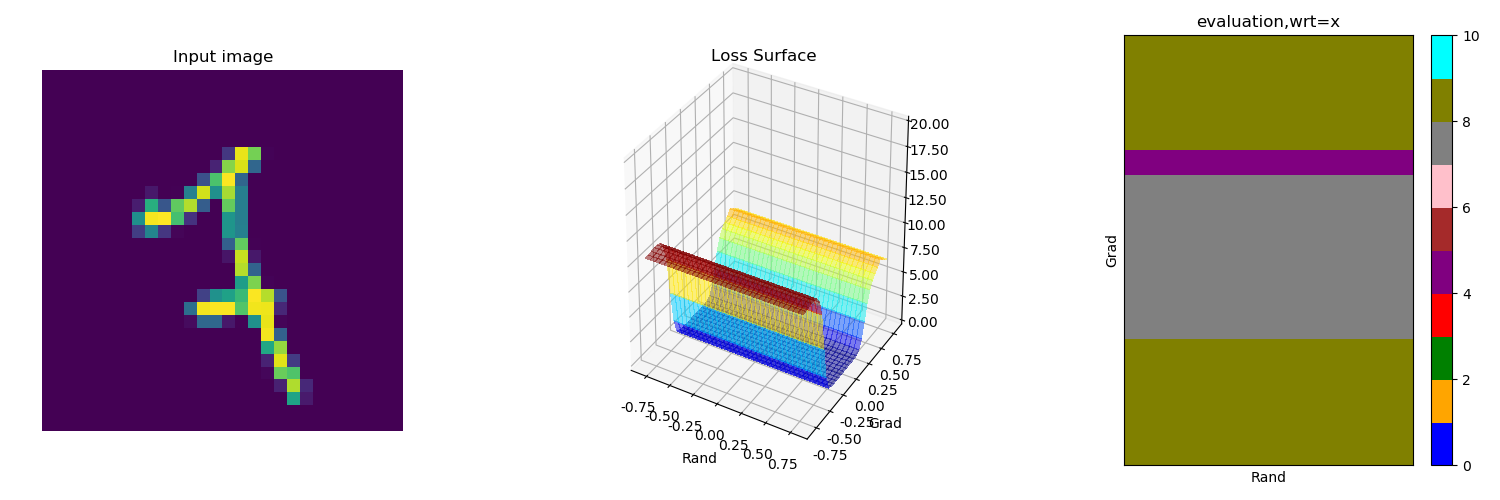}
\par\end{centering}
\begin{centering}
\includegraphics[width=0.35\textwidth]{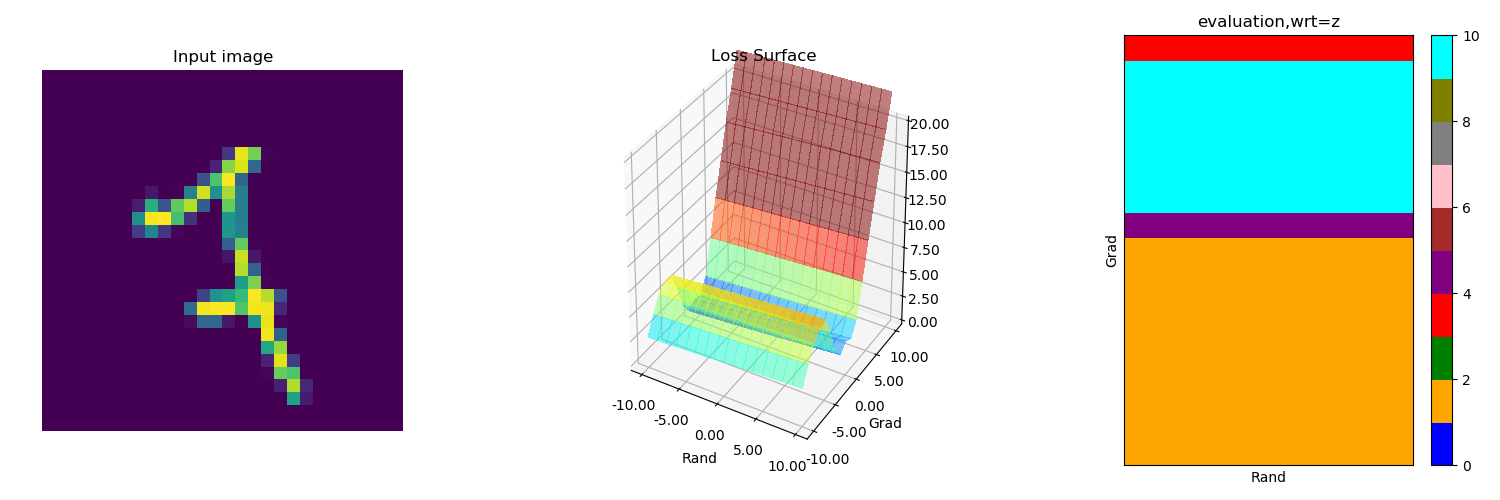}~~\includegraphics[width=0.35\textwidth]{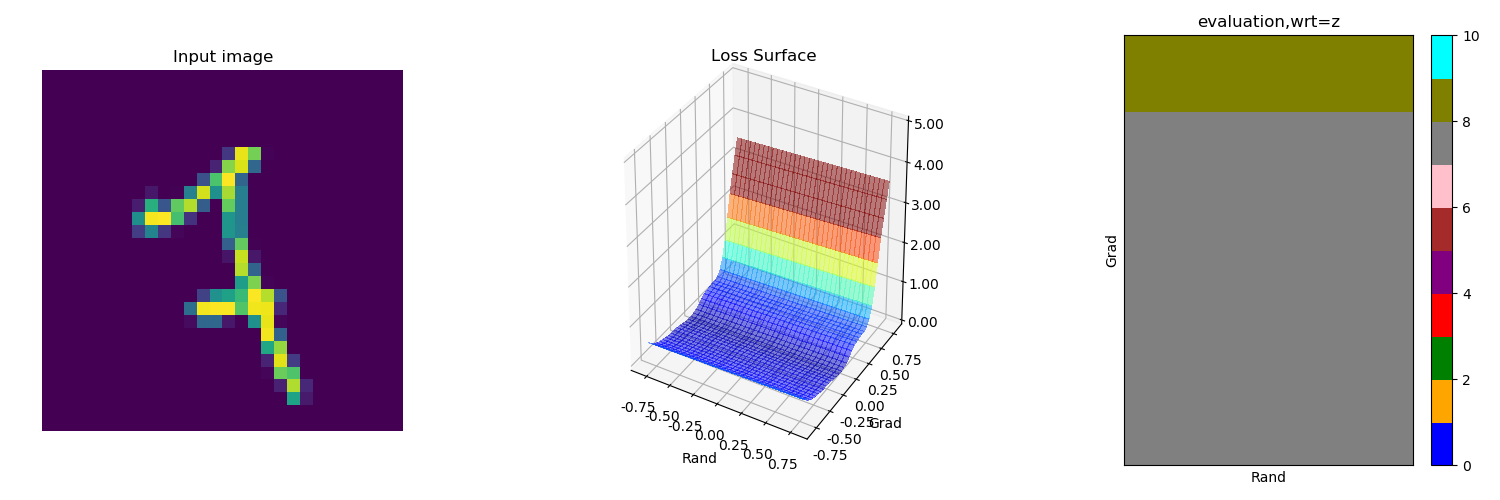}
\par\end{centering}
\caption{Loss surface at local region of a clean data example. Top-left: ADR-None
w.r.t input. Top-right: ADR+LC w.r.t input. Bottom-left: ADR-None
w.r.t latent. Bottom-right: ADR+LC w.r.t latent\label{fig:pred-surface-clean}}
\end{figure*}

To gain a better understanding of the contribution of the local compactness
component, we visualize the loss surface of the base model (ADV as
ADR-None) and the base model with only the local compactness term
(ADR+LC). In Figure \ref{fig:pred-surface-clean}, the left image
is a clean data example $x$, while the middle image is the loss surface
over the input region around $x$ in which the z-axis indicates the
cross-entropy loss w.r.t.$~$the true label (the higher value means
more incorrect prediction) and the x- and y-axis indicates the variance
of the input image along the gradient direction w.r.t.$~$$x$ and
a random orthogonal direction, respectively. By varying along the
two axes, we create a grid of images which represents the neighborhood
region around $x$. The right-hand image depicts the predicted labels
corresponding with this input grid. 

From Figure \ref{fig:pred-surface-clean}, for ADR-None, that its
neighborhood region is non-smooth, resulting in incorrect predictions
to the label 1 and 4. Meanwhile, for our ADR+LC method (adversarial
training with local compactness), the loss surface w.r.t.$~$the input
is smoother in its neighborhood region, resulting in correct predictions.
In addition, in our method, the prediction surface w.r.t.$~$the latent
feature in the intermediate representation layer is smoother than
that w.r.t.$~$input. This means that our local compactness makes
the local region more compact, hence improving adversarial robustness.
Visualization with an adversarial example as input can be found in
our supplementary material which provides more evidence of our improvement
over the base model. 

\begin{figure}
\begin{centering}
\includegraphics[width=0.4\textwidth]{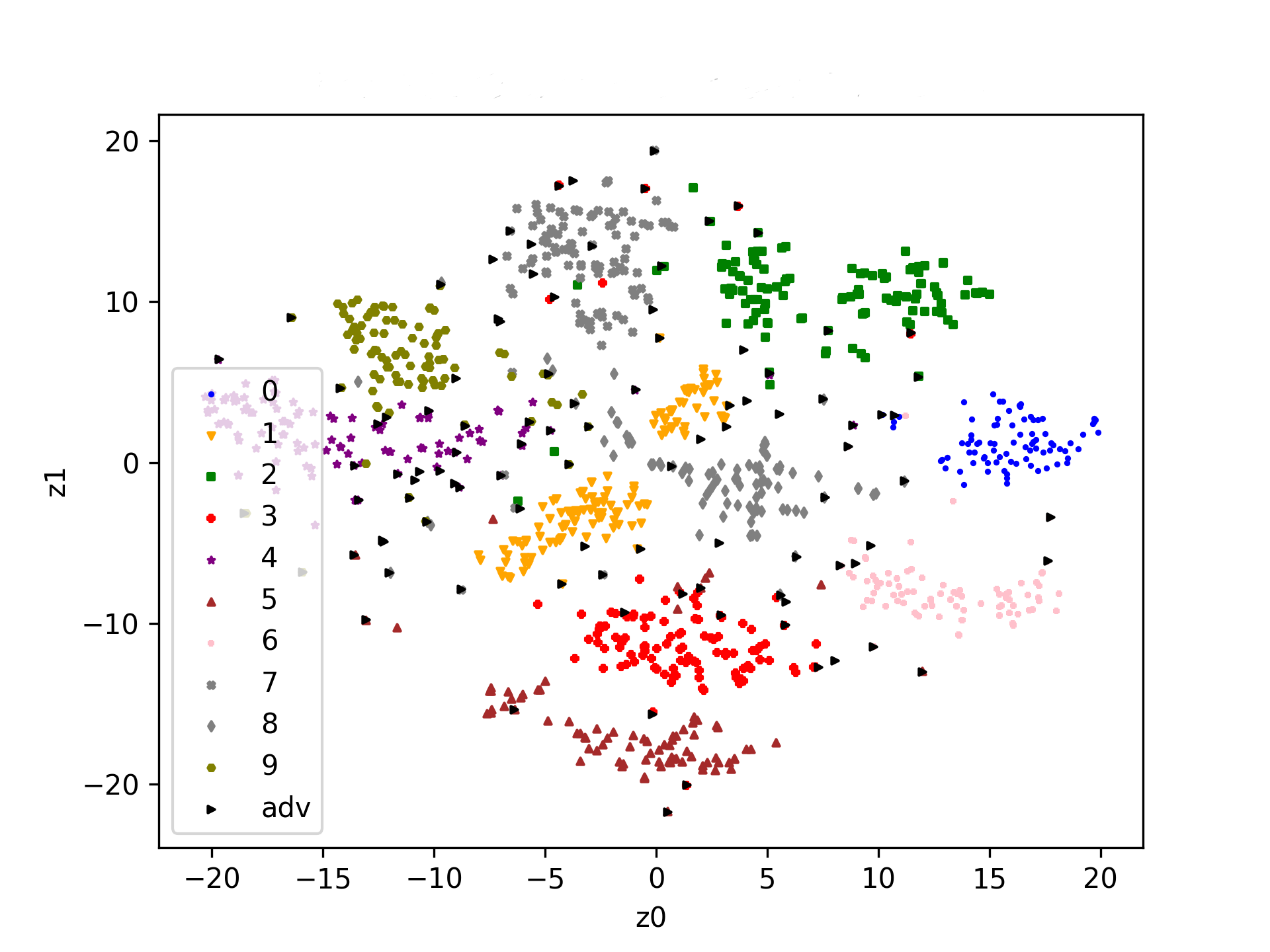}~~\includegraphics[width=0.4\textwidth]{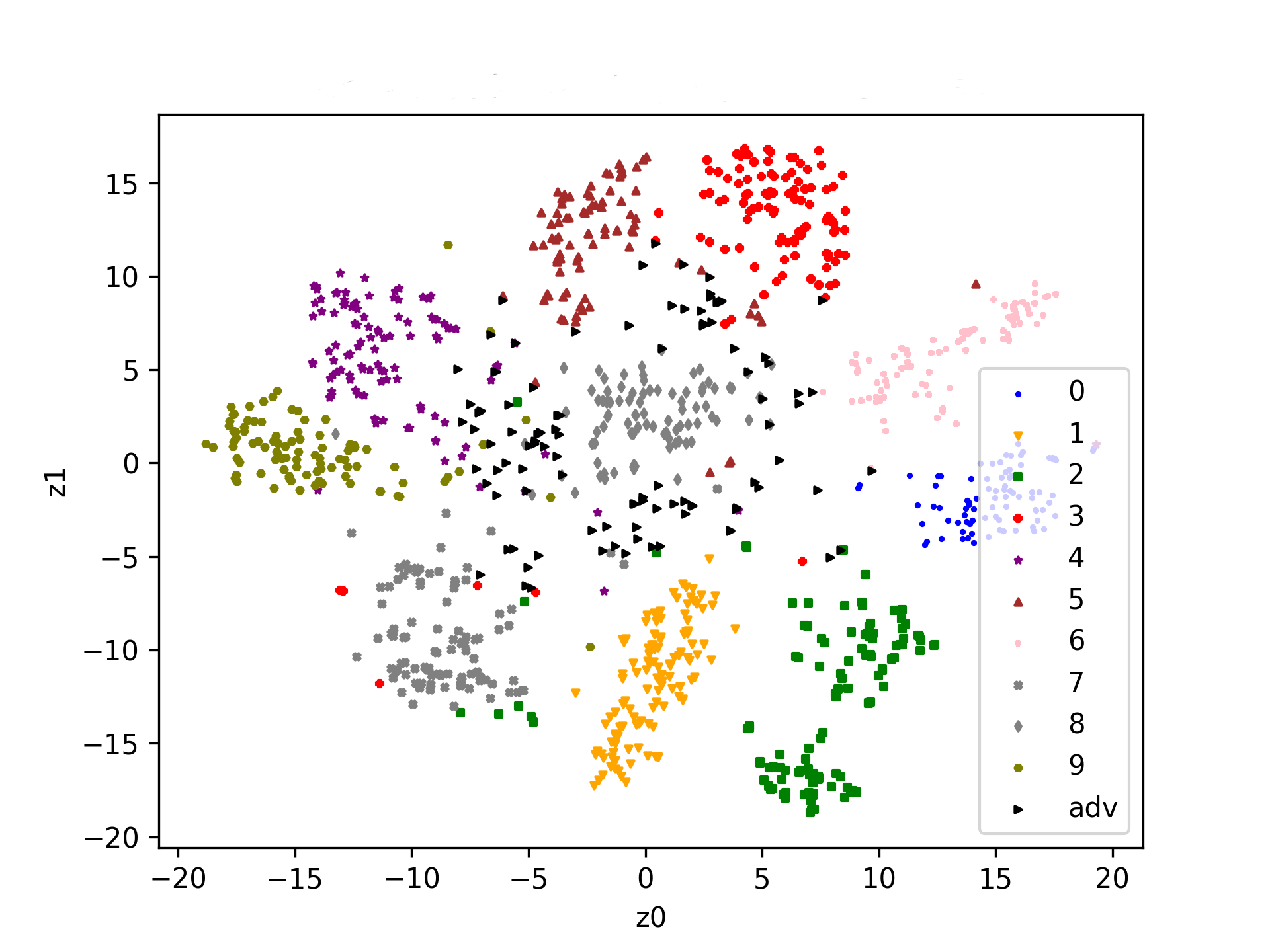}
\par\end{centering}
\caption{T-SNE visualization of latent space. Black triangles are (positive)
adversarial examples while others are clean images. Left: ADR-None.
Right: ADR+LC/GB\label{fig:tsne}}

\end{figure}

\begin{figure}
\begin{centering}
\includegraphics[width=0.4\textwidth]{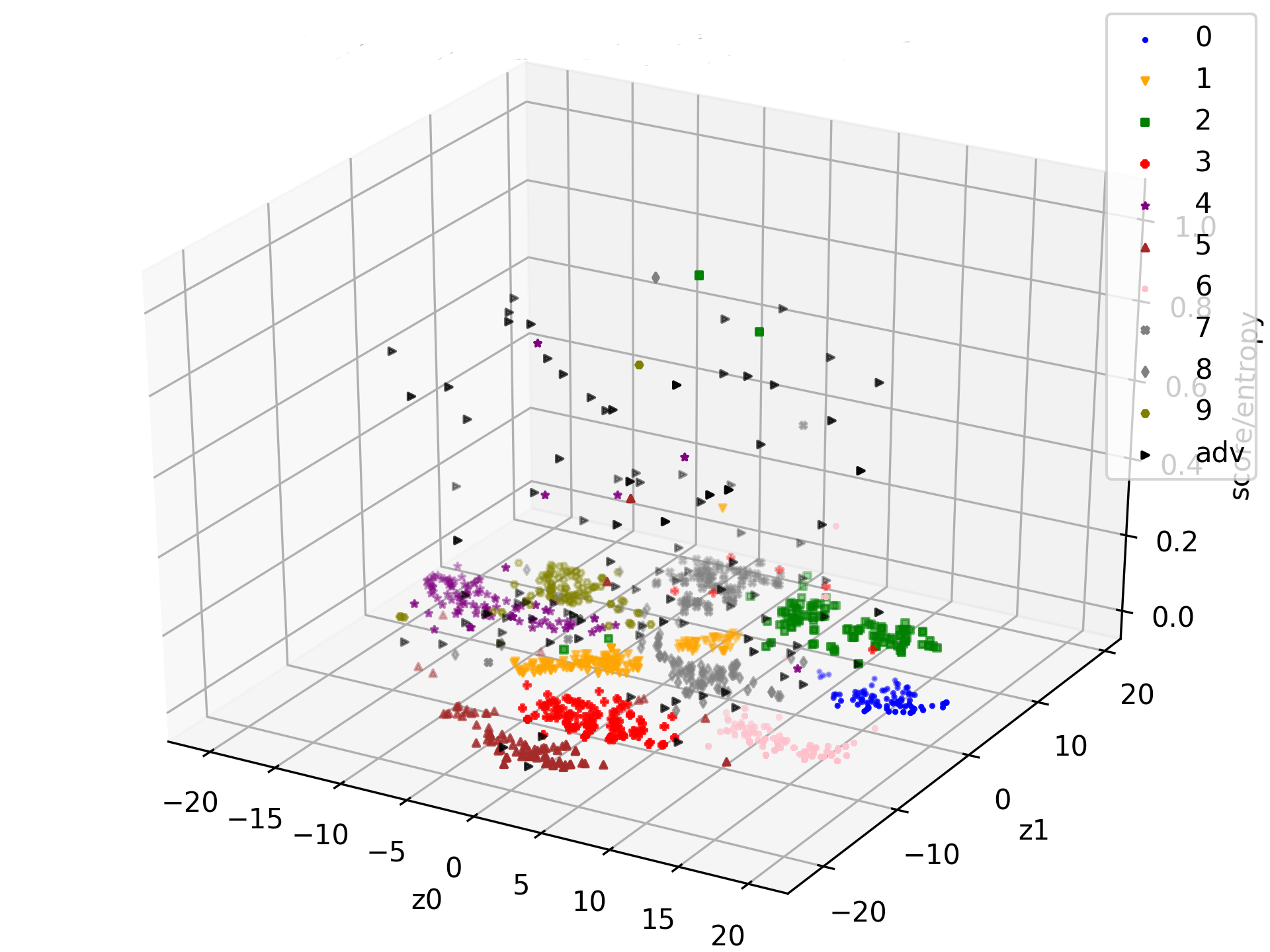}~~\includegraphics[width=0.4\textwidth]{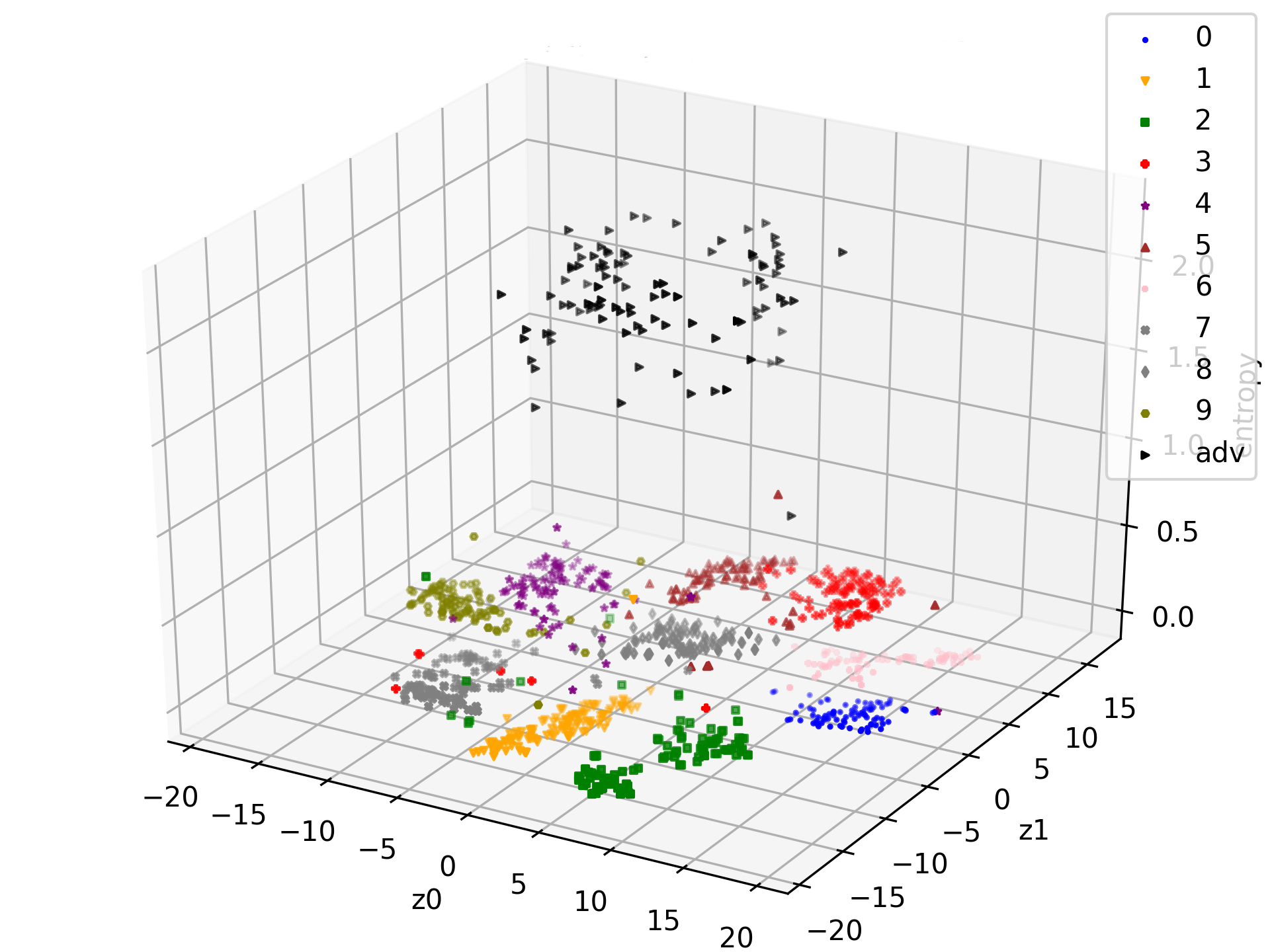}
\par\end{centering}
\caption{T-SNE visualization of latent space with entropy of the prediction
probability. Black triangles are (positive) adversarial examples while
others are clean images. Left: ADR-None. Right: ADR+LC/GB\label{fig:tsne-entropy}}

\end{figure}

Furthermore, we use t-SNE \cite{maaten2008visualizing} to visualize
the intermediate space for demonstrating the effect of our global
compactness component. We choose to show a \textit{positive adversarial example}
defined as an adversarial example which successfully fools a defense
method. We compare the base model (ADV as ADR-None) with our method
with the compactness terms and use t-SNE to project clean data and
adversarial examples onto 2D space as in Figure$~$\ref{fig:tsne}.
For ADR-None, its adversarial examples seem to distribute more broadly
and randomly. With our global compactness constraint, the adversarial
examples look well-clustered in a low density region, while rarely
present in the high density region of natural clean images. We leverage
the entropy of the prediction probability of examples as the third
dimension in Figure$~$\ref{fig:tsne-entropy}. A lower entropy mean
that the prediction is more confident (i.e., closer to a one-hot vector)
and vice versa. It can be observed that for the base model, the prediction
outputs of adversarial examples seem to be randomly distributed, while
for our ADR+LC/GB method, the prediction outputs of adversarial examples
mainly lie in the high entropy region and are well-separated from
those of the clean data examples. In other words, adversarial examples
can be more easily detected from clean examples in our method, according
to the predication entropy. In addition, the visualization for a \textit{negative adversarial example}
can be found in our supplementary material.


To summarize, in this ablation study, we have demonstrated how our
proposed components can improve adversarial robustness. In the next
section, we will compare the best variant (with all components) of
our method with both ADV and TRADES on more complex datasets to highlight
the capability of our method. 

\subsubsection{Experiment on the MNIST dataset}

\begin{figure*}[t]
\begin{centering}
\includegraphics[width=0.45\textwidth]{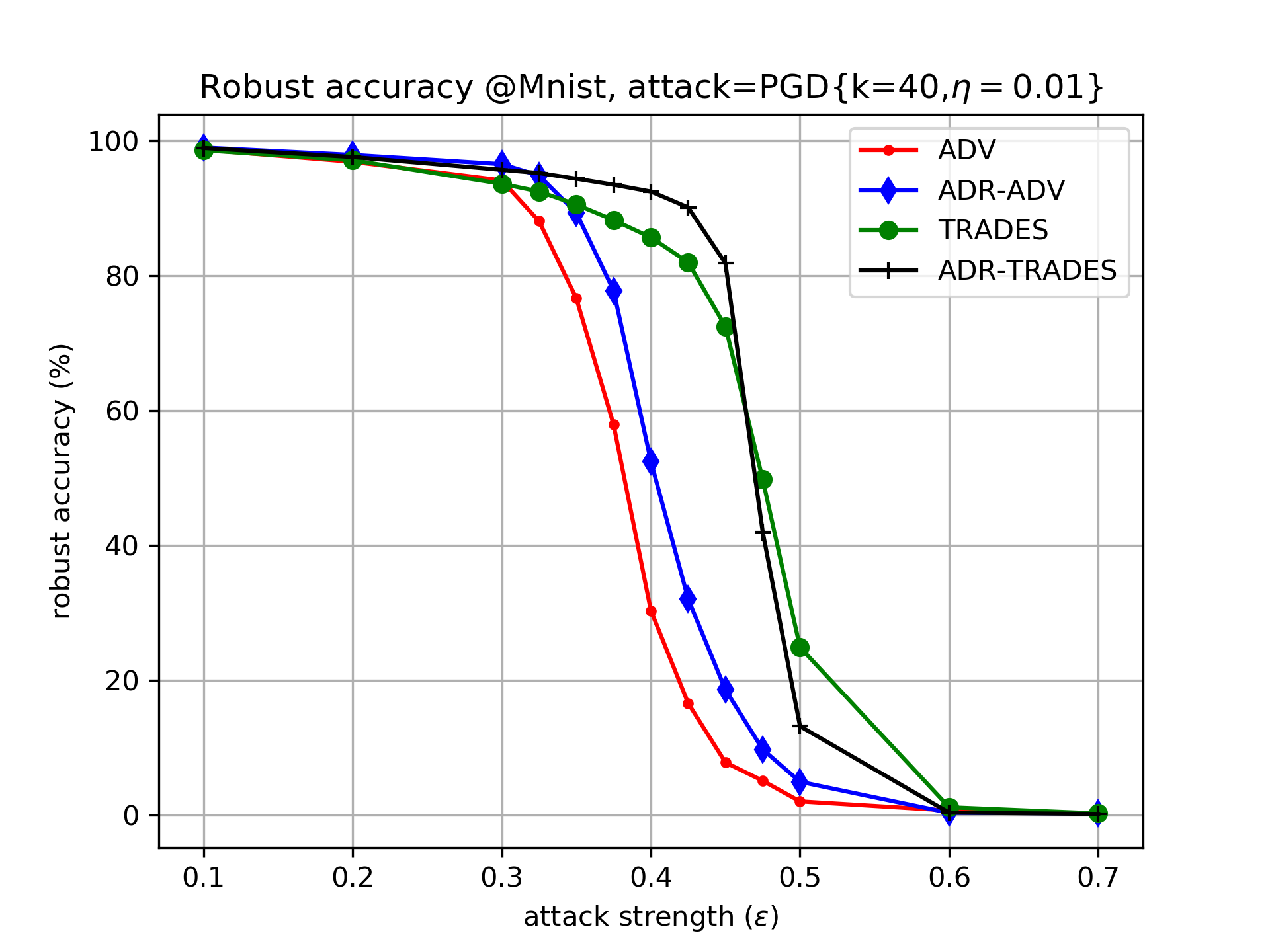}~~\includegraphics[width=0.45\textwidth]{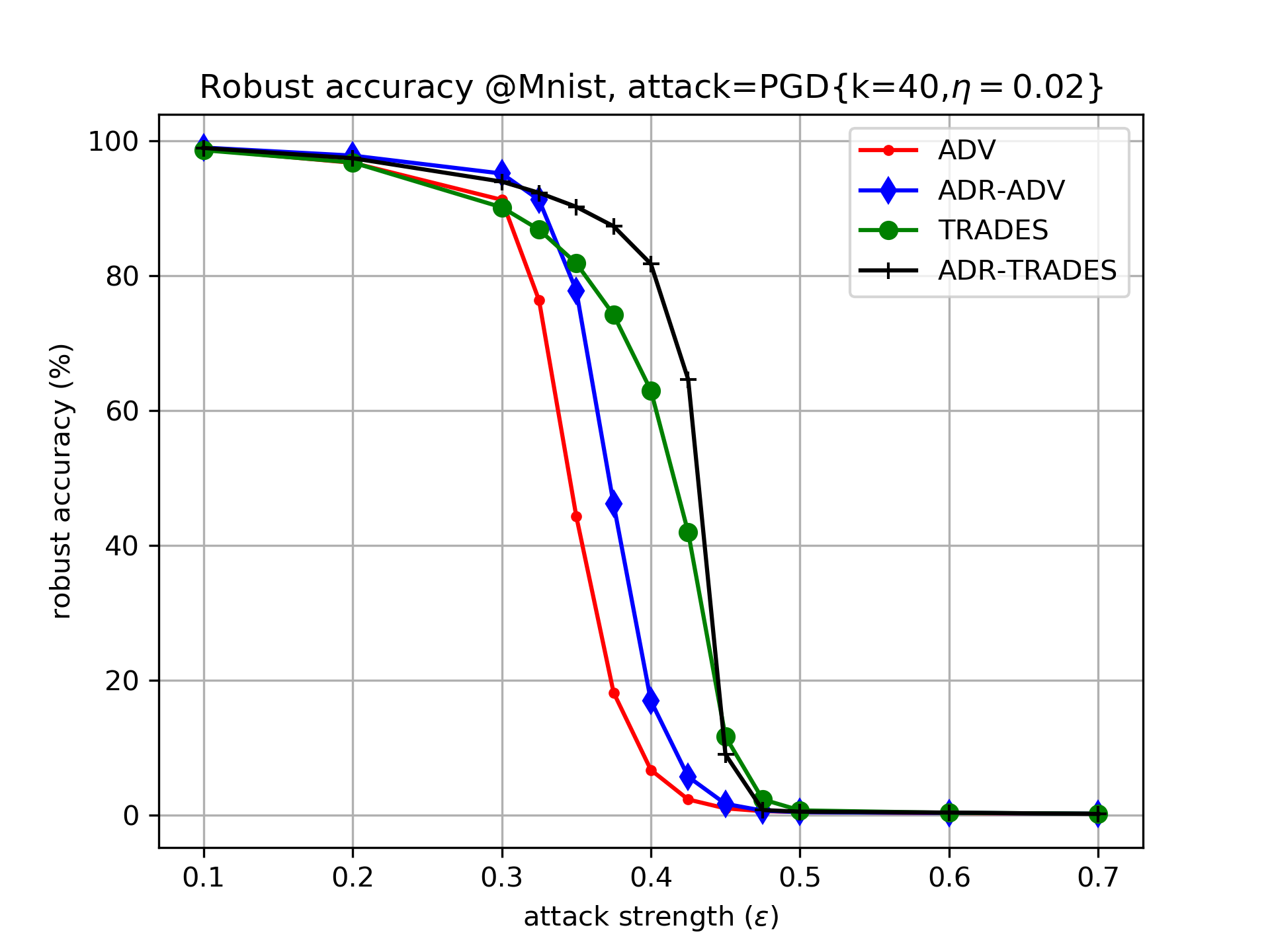}
\par\end{centering}
\caption{Robust accuracy against PGD attack on MNIST. Base models include ADV
and TRADES. Left: $\eta=0.01$. Right: $\eta=0.02$\label{fig:compare-mnist-acc}}
\end{figure*}

We compare our method with ADV and TRADES on the MNIST dataset. For
our method, in addition to using its full version with all of the
proposed terms, we consider two variants ADR-ADV and ADR-TRADES wherein
the adversary $\mathcal{A}$ is set to be ADV and TRADES respectively.
We use PGD/TRADES generated adversarial examples with $k=20,\varepsilon_{d}=0.3,\eta_{d}=0.01$
for adversarial training as proposed in \cite{madry2017towards} and
employ the PGD attack with $k=40$, using two iterative size $\eta\in\{0.01,0.02\}$
and different distortion boundaries $\varepsilon$ to attack. The
results shown in Figure \ref{fig:compare-mnist-acc} illustrate that
our variants outperform the baselines, especially for $\{\varepsilon=\varepsilon_{d}=0.3,\eta=0.01\}$.
For example, our ADR-ADV improves ADV by 2.4\% (from 94.15\% to 96.55\%)
while ADR-TRADES boosts TRADES by 2.07\% (from 93.64\% to 95.71\%).
While for attack setting $\{\epsilon=\epsilon_{d}=0.3,\eta=0.02\}$,
our method improves ADV and TRADES by 4.0\% and 3.8\% respectively.
Moreover, the improvement gap increases when the attack goes stronger.

\subsubsection{Experiment on the CIFAR-10 dataset}

\begin{figure*}[t]
\begin{centering}
\includegraphics[width=0.45\textwidth]{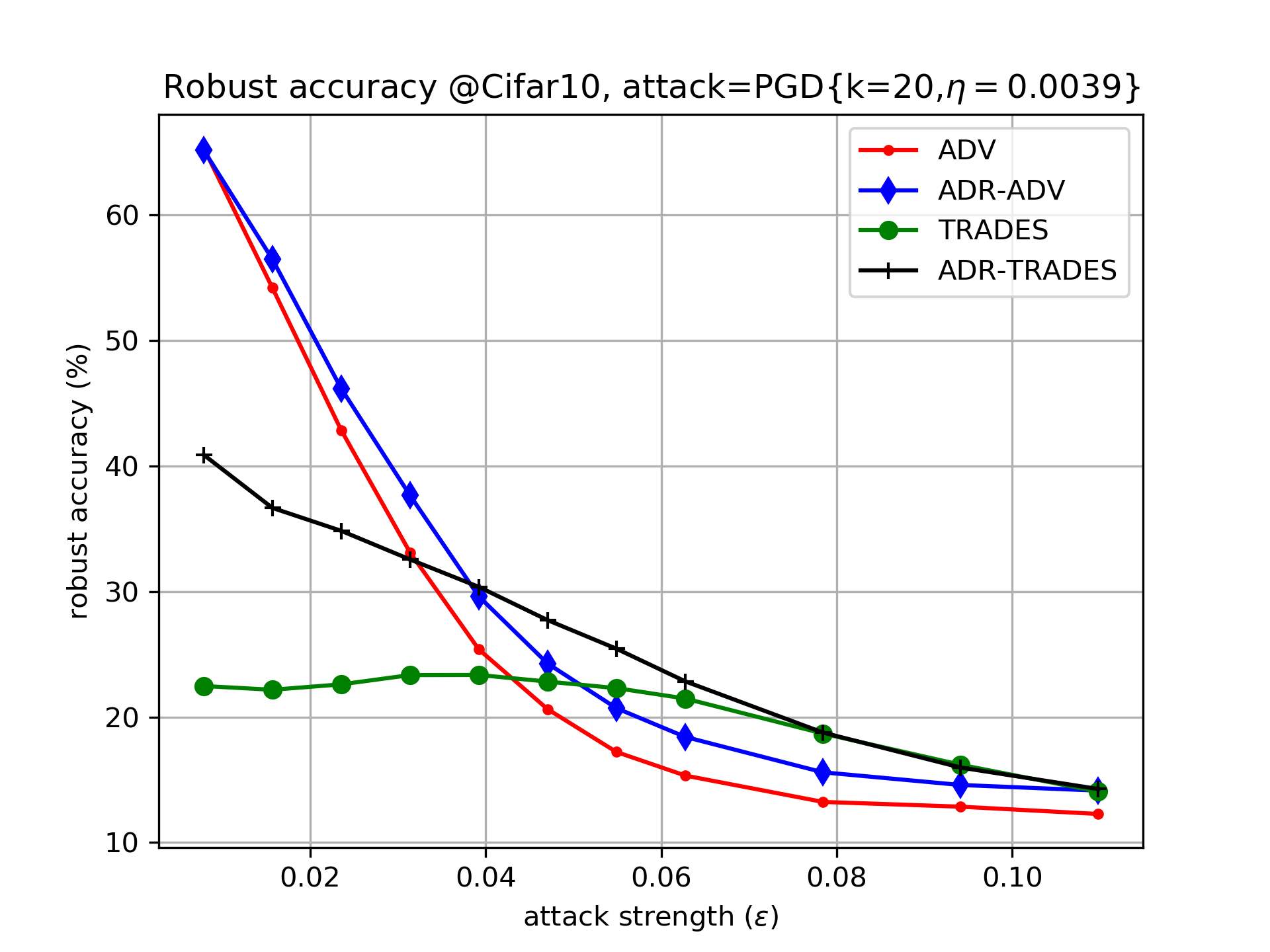}~~\includegraphics[width=0.45\textwidth]{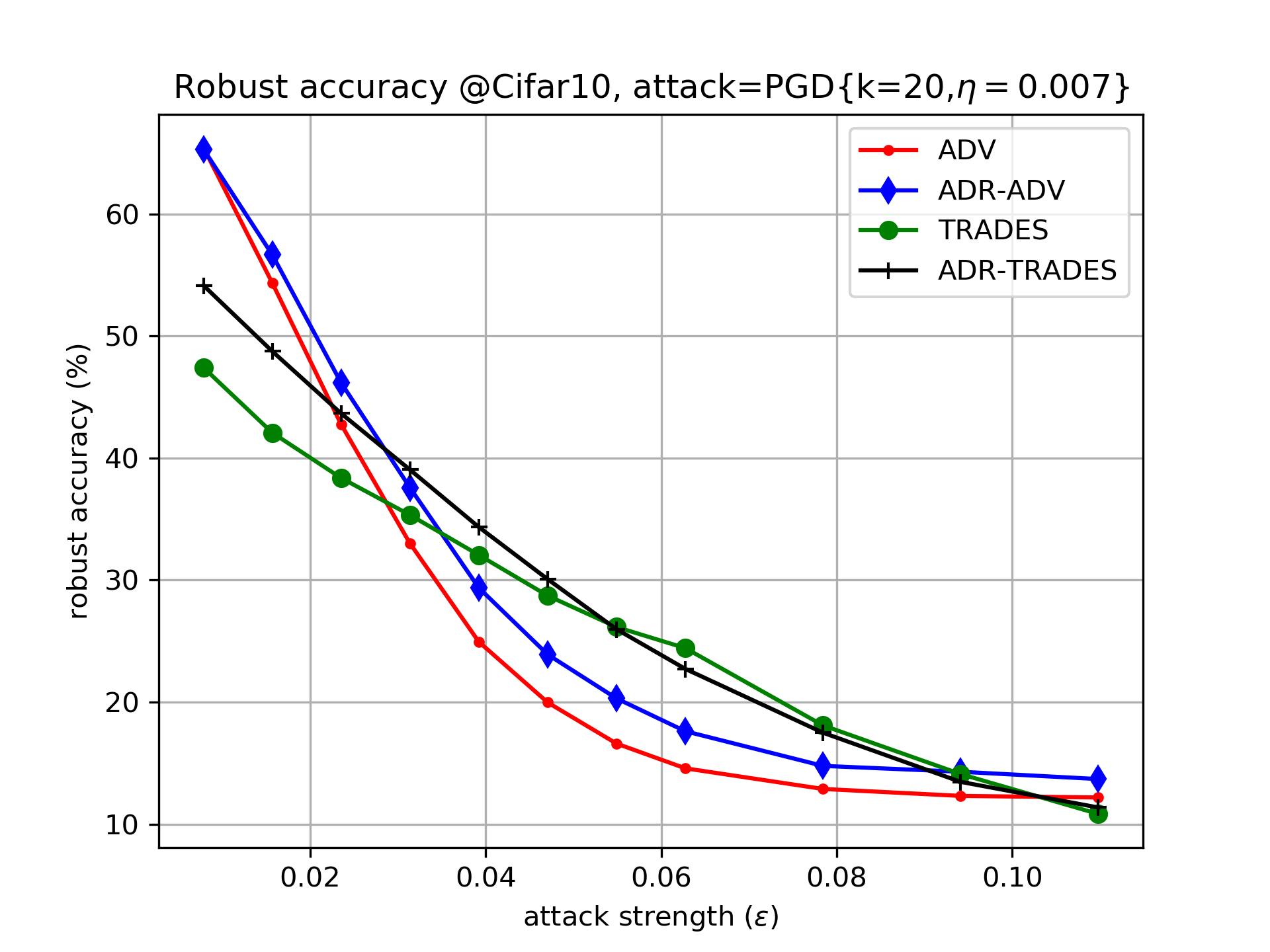}
\par\end{centering}
\caption{Robust accuracy against PGD attack on CIFAR-10, using Standard CNN
architecture. Base models include ADV and TRADES. Left: $\eta=0.0039$.
Right: $\eta=0.007$.\label{fig:compare-cifar-cnn}}
\end{figure*}

We conduct experiments on the CIFAR-10 dataset under two different
architectures: standard CNN from \cite{carlini2017towards} and ResNet
from \cite{madry2017towards}. We set $k=10,\varepsilon_{d}=0.031,\eta_{d}=0.007$
for ADV and TRADES and use a PGD attack with $k=20$, $\eta\in\{0.0039,0.007\}$
and different distortion boundary $\varepsilon$. The results for
standard CNN architecture in figure \ref{fig:compare-cifar-cnn} show
that our methods significantly improve over the baselines. Moreover,
the results for standard CNN architecture at a checking point $\{\varepsilon=\varepsilon_{d}=0.031,\eta=\eta_{d}=0.007\}$
in Table \ref{tab:compare-cifar-cnn-resnet} show that our methods
significantly outperform their baselines in terms of natural and robust
accuracies. Moreover, Figure \ref{fig:compare-cifar-cnn} indicates
that our proposed methods can defend better in a wide range of attack
strength. Particularly, when with varied distortion boundary $\varepsilon$
in $\left[0.02,0.1\right]$, our proposed methods always produce better
robust accuracies than its baselines. Finally, the results for ResNet
architecture in Table \ref{tab:compare-cifar-cnn-resnet} show a slight
improvement of our methods comparing with ADV but around $2.5\%$
improvement from TRADES on both Non-targeted and Multi-targeted attacks.\footnote{The performance of TRADES is influenced by the model architectures
and parameter tunings. The works \cite{qin2019adversarial,jalal2017robust}
also reported that TRADES cannot surpass ADV all the time which explains
the lower performance of TRADES on ResNet architecture in this paper.
More analysis can be found in the supplementary material.} The quality of adversarial examples and the chosen network architecture
obviously affects the overall performance, however, in this experiment,
we empirically prove that our proposed components can boost the robustness
under different combinations of the adversarial training frameworks
and network architectures.

\begin{table}
\begin{centering}
\caption{Robustness comparison on the CIFAR-10 dataset against PGD attack at
$k=20,\epsilon=0.031,\eta=0.007$ using Standard CNN and ResNet architectures\label{tab:compare-cifar-cnn-resnet}}
\par\end{centering}
\centering{}\resizebox{1.0\textwidth}{!}{
\begin{tabular}{cccc>{\centering}p{1mm}ccc}
\hline 
\multirow{2}{*}{} & \multicolumn{3}{c}{Standard CNN} &  & \multicolumn{3}{c}{ResNet}\tabularnewline
\cline{2-4} \cline{3-4} \cline{4-4} \cline{6-8} \cline{7-8} \cline{8-8} 
 & Nat. acc. & Non-target & Mul-target &  & Nat. acc. & Non-target & Mul-target\tabularnewline
\hline 
Standard model & 75.27\% & 12.26\% & 0.00\% &  & 92.51\% & 0.00\% & 0.00\%\tabularnewline
ADV & 67.86\% & 33.12\% & 18.73\% &  & 78.84\% & 44.08\% & 41.20\%\tabularnewline
TRADES & 71.37\% & 35.84\% & 18.01\% &  & 83.27\% & 37.52\% & 35.05\%\tabularnewline
ADR-ADV & 69.09\% & 37.67\% & 22.58\% &  & 78.43\% & 44.72\% & 41.43\%\tabularnewline
ADR-TRADES & 69.0\% & 39.68\% & 26.7\% &  & 82.02\% & 40.17\% & 37.70\%\tabularnewline
\hline 
\end{tabular}}
\end{table}

\section{Conclusion}

\vspace{-3mm}

Previous studies have shown that adversarial training has been one
of the few defense models resilient to various attack types against
deep neural network models. In this paper, we have shown that by enforcing
additional components, namely local/global compactness constraints
together with the clustering assumption, we can further improve the
state-of-the-art adversarial training models. We have undertaken comprehensive
experiments to investigate the effect of each component and have demonstrated
the capability of our proposed methods in enhancing adversarial robustness
using real-world datasets.

\paragraph{\textbf{Acknowledgement}: }

This work was partially supported by the Australian Defence Science
and Technology (DST) Group under the Next Generation Technology Fund
(NTGF) scheme.

\bibliographystyle{splncs04}
\addcontentsline{toc}{section}{\refname}\bibliography{5686}

\clearpage{}

\begin{center}
\textbf{\Large{}Supplementary to ``Improving Adversarial Robustness
by Enforcing Local and Global Compactness''}{\Large\par}
\par\end{center}

\section{Hyperparameters }

The hyperparameters for our experiments as Table \ref{tab:hyperparams}.
The hyperparameters of local compactness, global compactness, and
smoothness are set to be either 1 or 0, meaning they are switched
ON/OFF. Although finer tuning of these parameters can lead to better
results, our method outperforms the baselines in these initial settings,
which demonstrates the effectiveness of those components.

\begin{table}
\caption{Hyper-parameter setting for the experiment section\label{tab:hyperparams}}

\centering{}%
\begin{tabular}{ccccc}
\hline 
 & $\lambda_{com}^{lc}$ & $\lambda_{com}^{gb}$ & $\lambda_{smt}$ & $\lambda_{conf}$\tabularnewline
\hline 
MNIST & 1. & 1. & 1. & 0.\tabularnewline
CIFAR-10-CNN & 1. & 1. & 1. & 1.\tabularnewline
CIFAR-10-ResNet & 1. & 1. & 1. & 0.\tabularnewline
\hline 
\end{tabular}
\end{table}

\section{Model architectures and experimental setting}

We summarize the experimental setting in Table$~$\ref{tab:setting}.

\begin{table*}[p]
\begin{centering}
\caption{Experimental settings for our experiments. The model architectures
are from \cite{carlini2017towards} \cite{madry2017towards} and redescribed
in the supplementary material. \label{tab:setting}}
\par\end{centering}
\centering{}\resizebox{1.0\textwidth}{!}{
\begin{tabular}{ccccccc}
\hline 
 &  & MNIST &  & CIFAR-10 (CNN) &  & CIFAR-10 (Resnet)\tabularnewline
\hline 
Architectures &  & CNN-4C3F(32)\cite{carlini2017towards} &  & CNN-4C3F(64)\cite{carlini2017towards} &  & RN-34-10\cite{madry2017towards}\tabularnewline
Optimizer &  & SGD &  & Adam &  & SGD\tabularnewline
Learning rate &  & 0.01 &  & 0.001 &  & 0.1\tabularnewline
Momentum &  & 0.9 &  & N/A &  & 0.9\tabularnewline
Training stratery &  & Batch size 128, 100 epochs &  & Batch size 128, 200 epochs &  & Batch size 128, 200 epochs\tabularnewline
Perturbation &  & $k=20,\epsilon_{d}=0.3,\eta_{d}=0.01,l_{\infty}$ &  & $k=10,\epsilon_{d}=0.031,\eta_{d}=0.007,l_{\infty}$ &  & $k=10,\epsilon_{d}=0.031,\eta_{d}=0.007,l_{\infty}$\tabularnewline
\hline 
\end{tabular}}\vspace{0mm}
\end{table*}

For the MNIST dataset, we used the standard CNN architecture with
three convolution layers and three fully connected layers described
in \cite{carlini2017towards}. For the CIFAR-10 dataset, we used two
architectures in which one is the standard CNN architecture described
in \cite{carlini2017towards} and another is the ResNet architecture
used in \cite{madry2017towards}. The ResNet architecture has 5 residual
units with (16, 16, 32, 64) filters each. We choose the convolution
layers as the Generator and the last fully connected layers as the
Classifier for ResNet architecture. The standard CNN architectures
are redescribed as follow: 

\textbf{CNN-4C3F(32) Generator}: 2$\times$Conv(32)$\rightarrow$
Max Pooling$\rightarrow$ 2$\times$Conv(32)$\rightarrow$ Max Pooling$\rightarrow$
Flatten 

\textbf{CNN-4C3F(32) Classifier}: FC(200)$\rightarrow$ ReLU$\rightarrow$
Dropout(0.5)$\rightarrow$ FC(200)$\rightarrow$ ReLU$\rightarrow$
FC(10)$\rightarrow$ Softmax 

\textbf{CNN-4C3F(64) Generator}: 2$\times$Conv(64)$\rightarrow$
Max Pooling$\rightarrow$ 2$\times$Conv(64)$\rightarrow$ Max Pooling$\rightarrow$
Flatten 

\textbf{CNN-4C3F(64) Classifier}: FC(256)$\rightarrow$ ReLU$\rightarrow$
Dropout(0.5)$\rightarrow$ FC(256)$\rightarrow$ ReLU$\rightarrow$
FC(10)$\rightarrow$ Softmax 

\section{Choosing the intermediate layer }

The intermediate layer for enforcing compactness constraints immediately
follows on from the generator. We additionally conduct an ablation
study to investigate the importance of choosing the intermediate layer
and report natural accuracy and robust accuracy against non-targeted/multiple-targeted
attacks respectively. We use the standard CNN architecture (which
has 4 Convolution layers in Generator and 3 FC layers in Classifier),
with four additional variants corresponding to different choices of
the intermediate layer (right after the generator). We use PGD ($k=100,\epsilon=0.3,\eta=0.01$
for MNIST, $k=100,\epsilon=0.031,\eta=0.007$ for CIFAR-10) to evaluate
these models. It can be seen from the results as showing in Table
\ref{tab:ab-choice} that the performance slightly downgrades if choosing
shallower layers. The higher impact is expected on a larger architecture
(i.e., Resnet), which can be investigated in future. 

\begin{table}
\caption{Performance comparison on different choices of the intermediate layer.
The results in each setting are natural accuracy and robust accuracy
against non-targeted/multiple-targeted attacks respectively. \label{tab:ab-choice}}

\begin{centering}
\begin{tabular}{c>{\centering}p{2mm}c>{\centering}p{2mm}c}
\hline 
 &  & MNIST &  & CIFAR10\tabularnewline
\hline 
G=2Conv, C=2Conv+3FC  &  & 99.52/93.88/92.78  &  & 68.78/36.46/21.99 \tabularnewline
G=3Conv, C=1Conv+3FC  &  & 99.44/94.38/93.59  &  & 69.17/37.05/22.44 \tabularnewline
CNN (G=4Conv, C=3FC) &  & 99.48/95.06/94.26  &  & 69.08/37.06/22.44 \tabularnewline
G=4Conv+1FC, C=2FC &  & 99.51/94.38/93.47  &  & 69.39/37.31/22.87 \tabularnewline
G=4Conv+2FC, C=1FC  &  & 99.52/94.26/93.45  &  & 69.13/37.31/22.57 \tabularnewline
\hline 
\end{tabular}
\par\end{centering}
\end{table}

\section{The performance of TRADES }

TRADES aims to find the most divergent adversarial examples, while
ADV aims to find the worst-case examples to improve a model (see Sec.
2.2 in our paper for more detail). Hence theoretically, there is no
guarantee that TRADES outperforms ADV. In practice, the performance
of TRADES is influenced by the classifier architectures and parameter
tunings. The works \cite{qin2019adversarial,jalal2017robust} also
reported that TRADES cannot surpass ADV all the time (Table 1 and
footnote 8 in \cite{qin2019adversarial}, Table 1 in \cite{jalal2017robust}),
which is in line with the findings in our paper. 

\section{Further experiments}

We conduct an additional evaluation with further state-of-the-art
attack methods (e.g., the Basic Iterative Method - BIM \cite{kurakin2016a}
and the Momentum Iterative Method - MIM \cite{dong2018boosting})
to convince that our method indeed boots the robustness rather than
suffers the gradient obfuscation \cite{athalye2018obfuscated}. Three
attack methods PGD, BIM and MIM share the same setting, i.e., $\{k=100,\epsilon=0.3,\eta=0.01\}$
for MNIST and $\{k=100,\epsilon=0.031,\eta=0.007\}$ for CIFAR-10.
The result as in Table \ref{tab:3atks} show that our components can
improve the robustness of the baseline framework against all three
kind of attacks which again proves the efficacy of our method.

\begin{table}
\caption{Robustness comparison on the MNIST and CIFAR-10 datasets using Standard
CNN with higher attack iteration (i.e., $k=100$). The results in
each setting are natural accuracy and robust accuracy against non-targeted/multiple-targeted
attacks respectively.\label{tab:3atks}}

\begin{centering}
\begin{tabular}{c>{\centering}p{2mm}c>{\centering}p{2mm}c>{\centering}p{2mm}c}
\hline 
 &  & Dataset &  & ADV &  & ADR-ADV\tabularnewline
\hline 
PGD &  & MNIST &  & 99.43/93.13/92.09 &  & 99.48/95.06/94.26\tabularnewline
BIM &  & MNIST &  & 99.43/93.00/91.70 &  & 99.48/94.86/93.99\tabularnewline
MIM &  & MNIST &  & 99.43/94.05/92.63 &  & 99.48/95.41/94.56\tabularnewline
PGD &  & CIFAR-10 &  & 67.61/32.87/18.74 &  & 69.16/36.85/22.71\tabularnewline
BIM &  & CIFAR-10 &  & 67.61/32.89/18.71 &  & 69.16/36.82/22.69\tabularnewline
MIM &  & CIFAR-10 &  & 67.61/33.00/18.59 &  & 69.16/36.96/22.56\tabularnewline
\hline 
\end{tabular}
\par\end{centering}
\end{table}

\subsection{Loss surface of adversarial examples }

\begin{figure*}
\begin{centering}
\vspace{-2mm}
\includegraphics[width=0.5\textwidth]{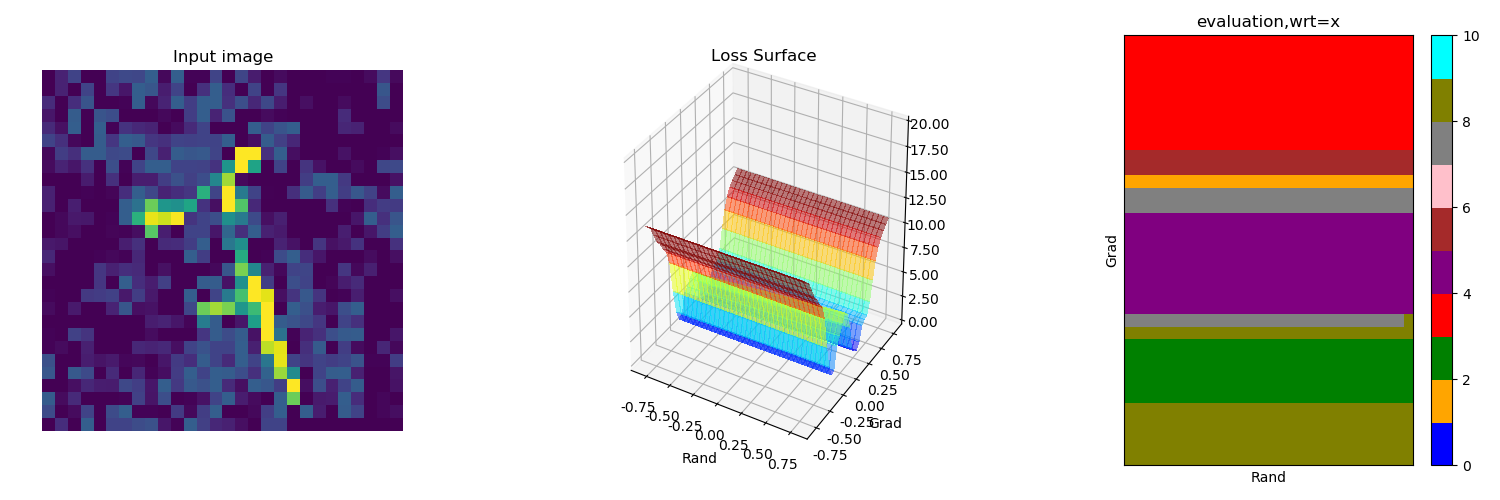}~~\includegraphics[width=0.5\textwidth]{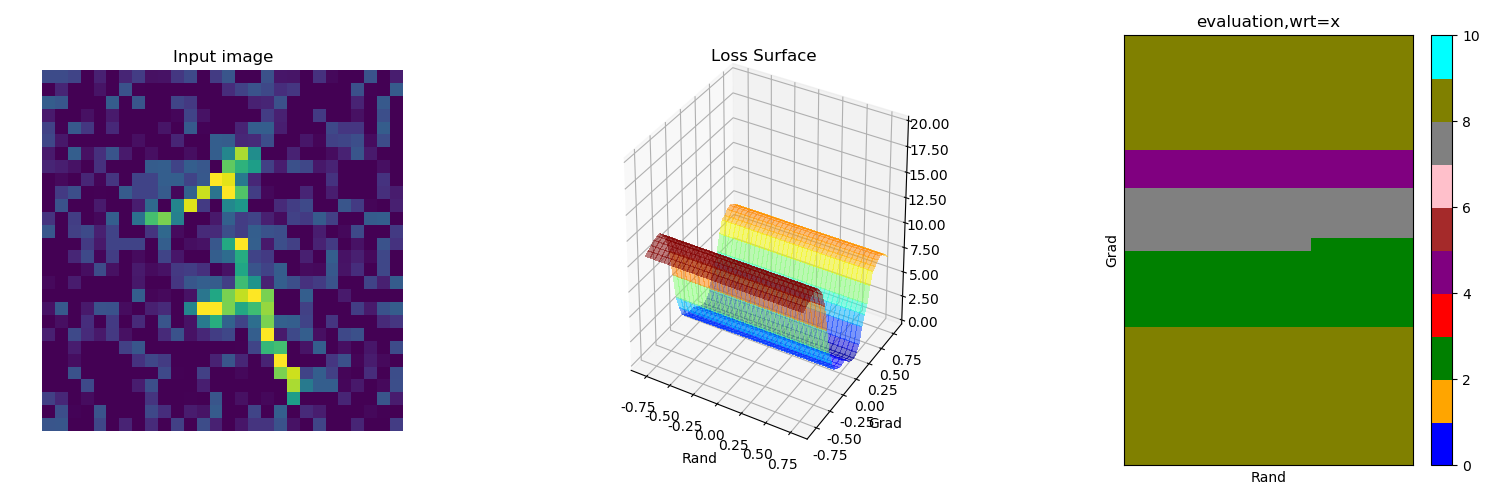}
\par\end{centering}
\begin{centering}
\includegraphics[width=0.5\textwidth]{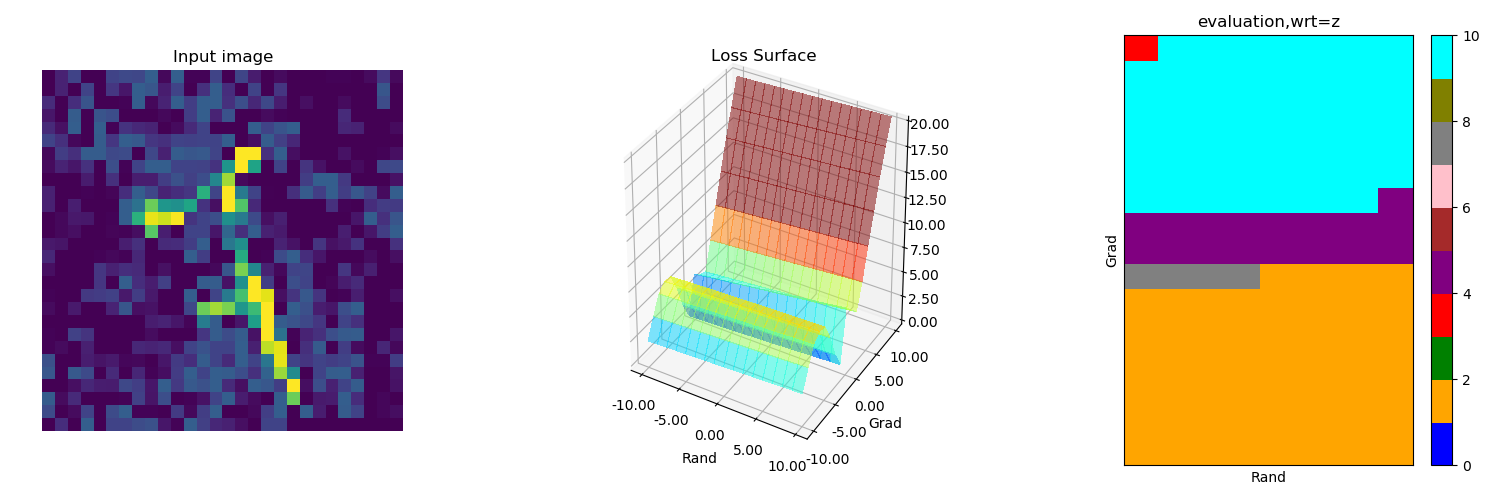}~~\includegraphics[width=0.5\textwidth]{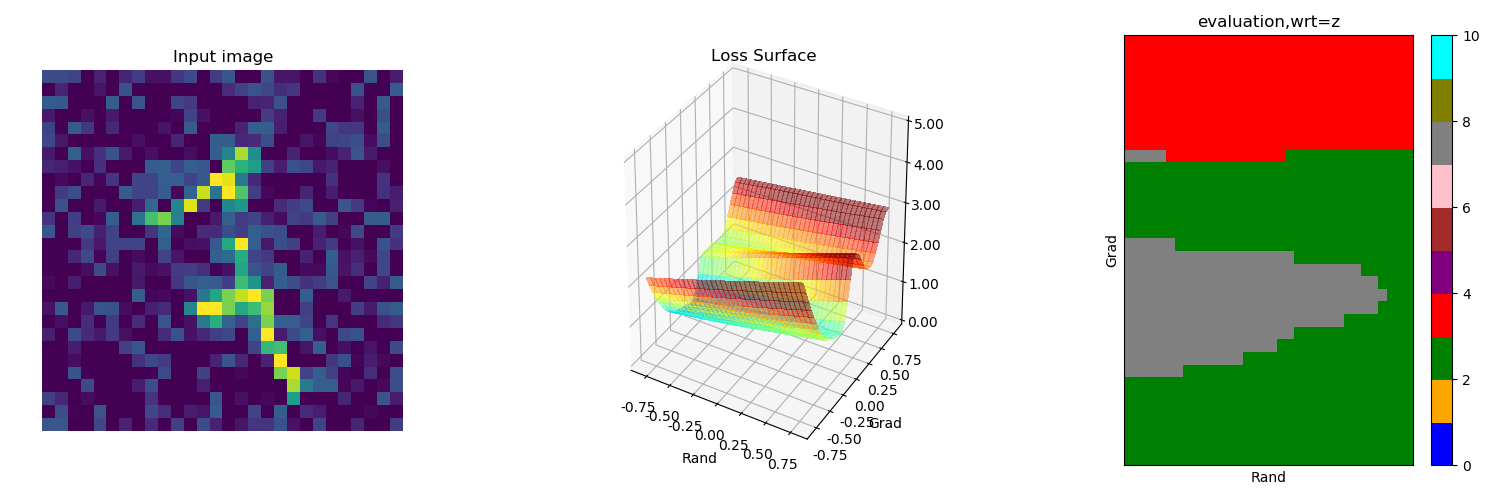}\vspace{-2mm}
\par\end{centering}
\caption{Loss surface at local region of a positive adversarial example. Top-left:
ADR-None w.r.t input. Top-right: ADV+LC w.r.t input. Bottom-left:
ADR-None w.r.t latent. Bottom-right: ADV+LC w.r.t latent\label{fig:pred-surface-supp}}
\vspace{-2mm}
\end{figure*}

We separate adversarial examples into two classes: positive adversarial
example which successfully fools a defense method and negative adversarial
example which is an unsuccessful attack. The loss surface of positive
adversarial example as Figure \ref{fig:pred-surface-supp}. In particular,
both ADV (ADR-None) and our method (ADR+LC) predicted $x_{a}$ with
the label 8, whereas its true label is 3. From Figure \ref{fig:pred-surface-supp},
it is evident that for ADV, that most of its neighborhood region is
non-smooth, resulting in incorrect predictions in almost all of the
grid. By contrast, for our method (ADR+LC), the loss surface w.r.t.$~$the
input is smoother, resulting in more correct predictions in this neighborhood
region. In addition, in our method, the prediction surface w.r.t.$~$the
latent feature in the intermediate representation layer is smoother
than that w.r.t.$~$input. This means that our local compactness makes
the local region more compact, hence improving adversarial robustness.

\begin{figure*}
\begin{centering}
\vspace{-2mm}
\includegraphics[width=0.5\textwidth]{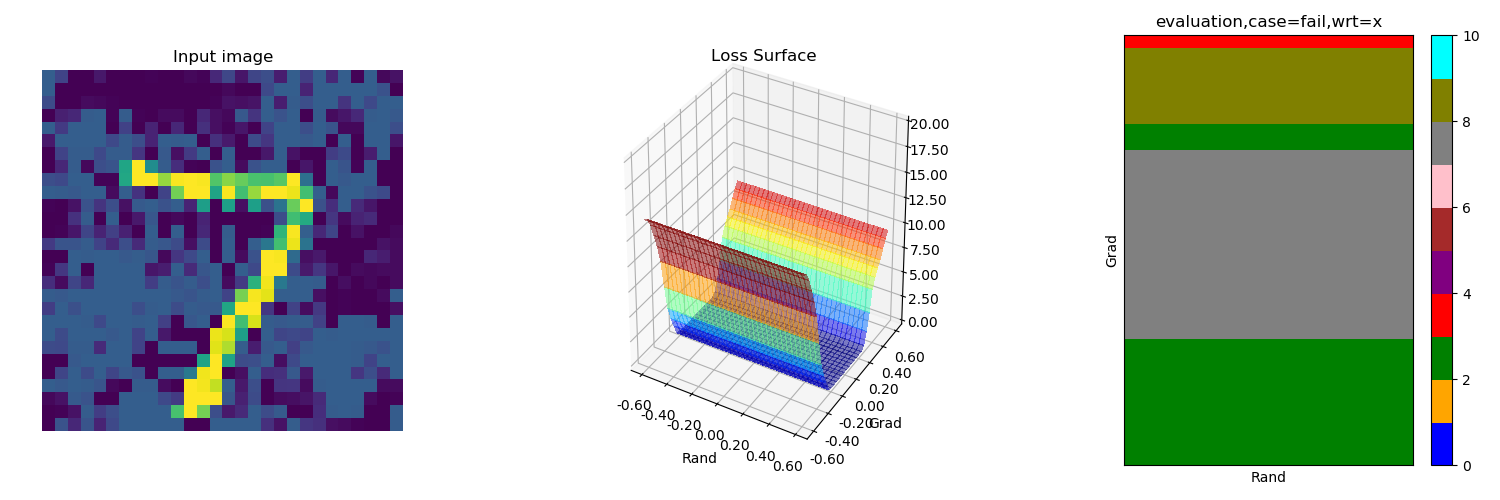}~~\includegraphics[width=0.5\textwidth]{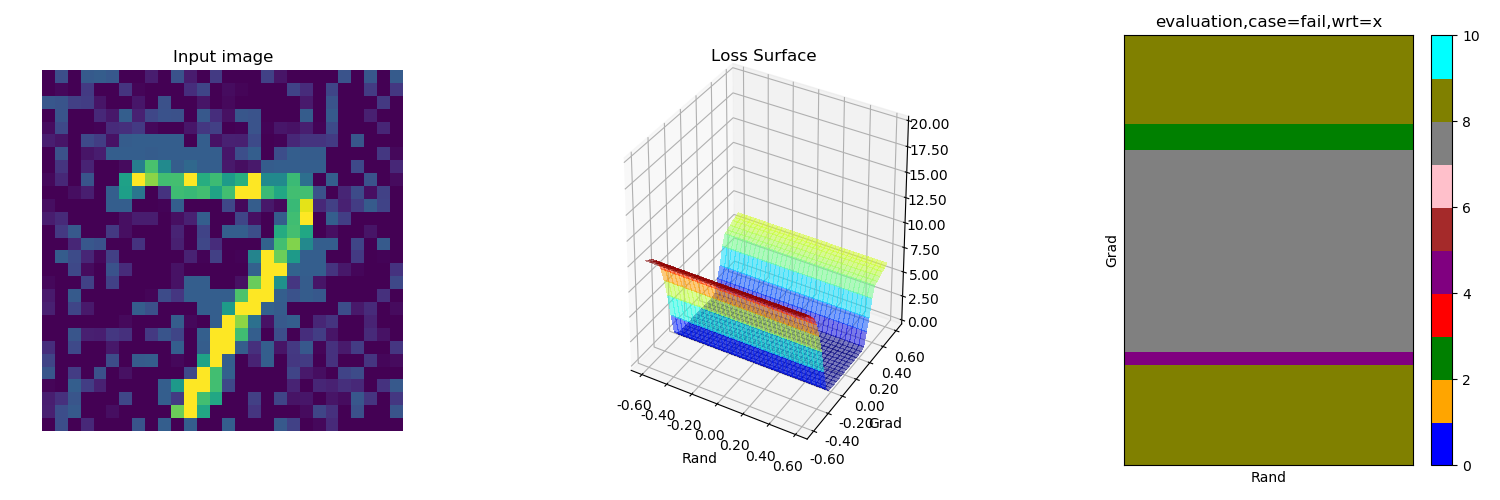}
\par\end{centering}
\begin{centering}
\includegraphics[width=0.5\textwidth]{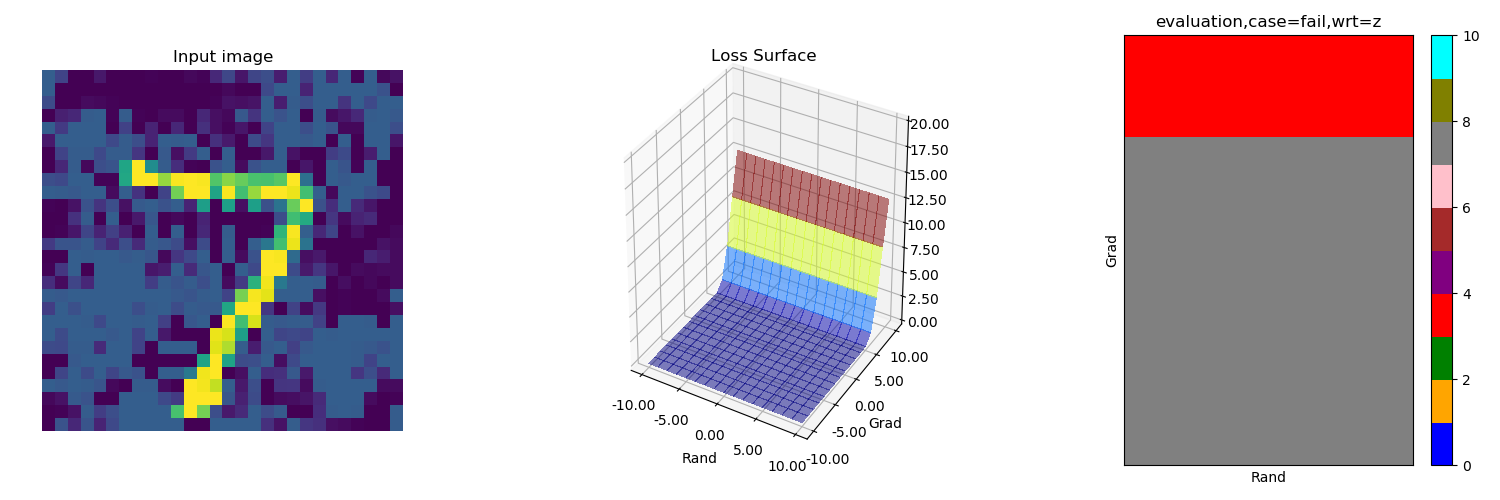}~~\includegraphics[width=0.5\textwidth]{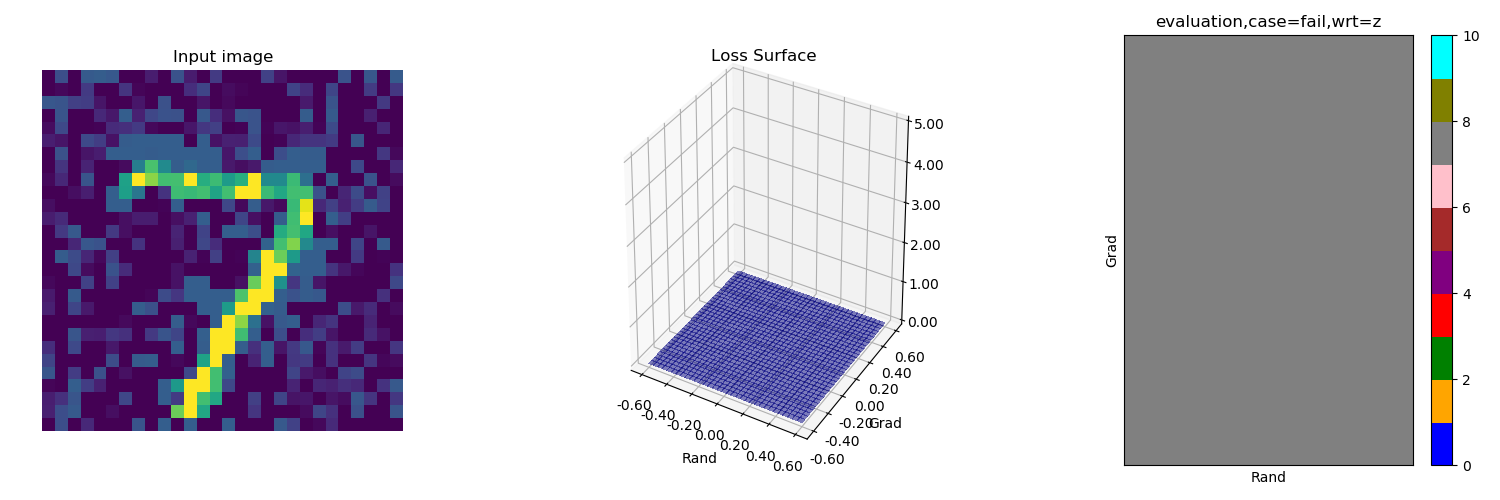}\vspace{-2mm}
\par\end{centering}
\caption{Loss surface at local region of a negative adversarial example. Top-left:
ADR-None w.r.t input. Top-right: ADV+LC w.r.t input. Bottom-left:
ADR-None w.r.t latent. Bottom-right: ADV+LC w.r.t latent\label{fig:surface-negative}}
\vspace{-2mm}
\end{figure*}

We provide the loss surface of negative adversarial examples from
adversarial training method and adversarial training with our components
as Figure \ref{fig:surface-negative}. Both examples show that the
loss function smooth in local region of an adversarial example. 

\subsection{T-SNE visualization of adversarial examples }

\begin{figure*}
\begin{centering}
\includegraphics[width=0.45\textwidth]{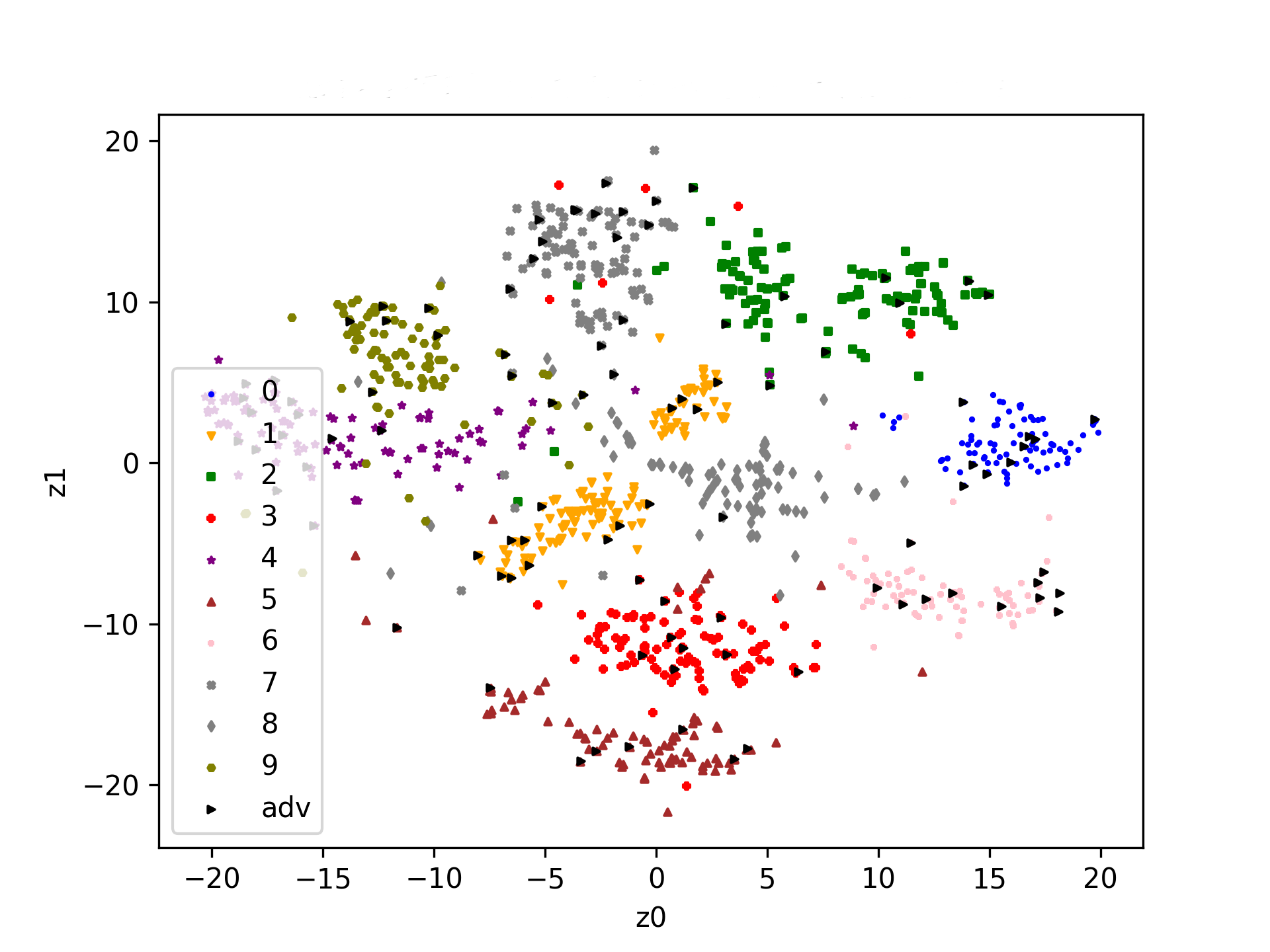}~~\includegraphics[width=0.45\textwidth]{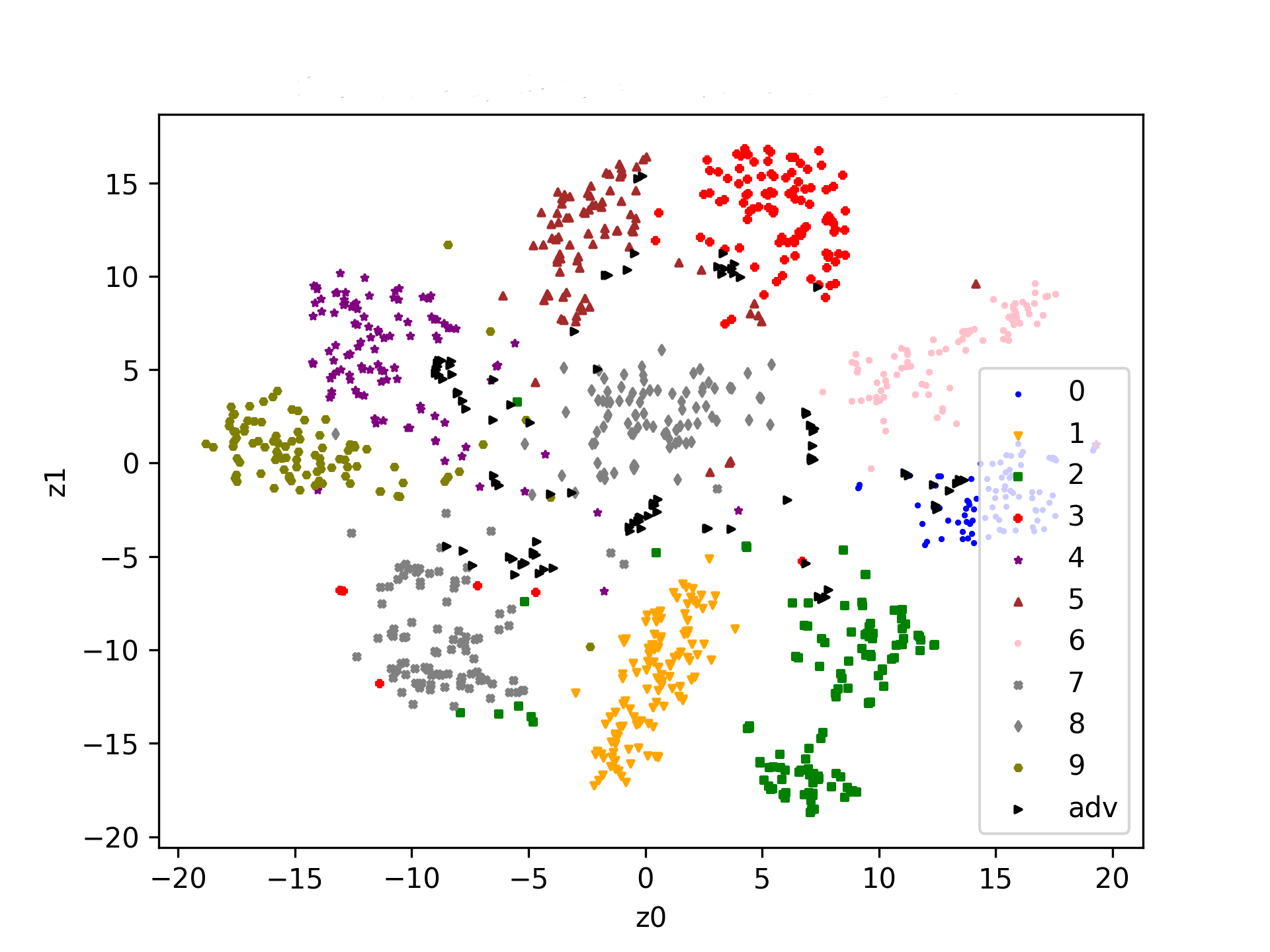}
\par\end{centering}
\caption{T-SNE visualization of latent space. Black triangles are (negative)
adversarial examples while others are clean images. Left: ADR-None.
Right: ADR+LC/GB\label{fig:neg-tsne}}
\end{figure*}

In addition to positive adversarial examples, we provide the t-SNE
visualization of the negative adversarial examples from adversarial
training (ADR-None) and adversarial training with our components (ADR+LC/GB)
as Figure \ref{fig:neg-tsne}. In adversarial training method, the
unsuccessful attacks have been mixed insight the natural/clean data.
In contrast, in case adversarial training with our components, the
attack representation consistently is separated from those from natural
data, similar to positive adversarial examples. Additionally, the
unsuccessful attacks in adversarial training have the same confidence
level with natural data, while those in our methods are totally different
levels. In summary, our method can produce a better latent representation
which is well separated between natural data and adversarial example
(both positive and negative). This feature can be used for adversarial
detection. 

\begin{figure*}
\begin{centering}
\includegraphics[width=0.45\textwidth]{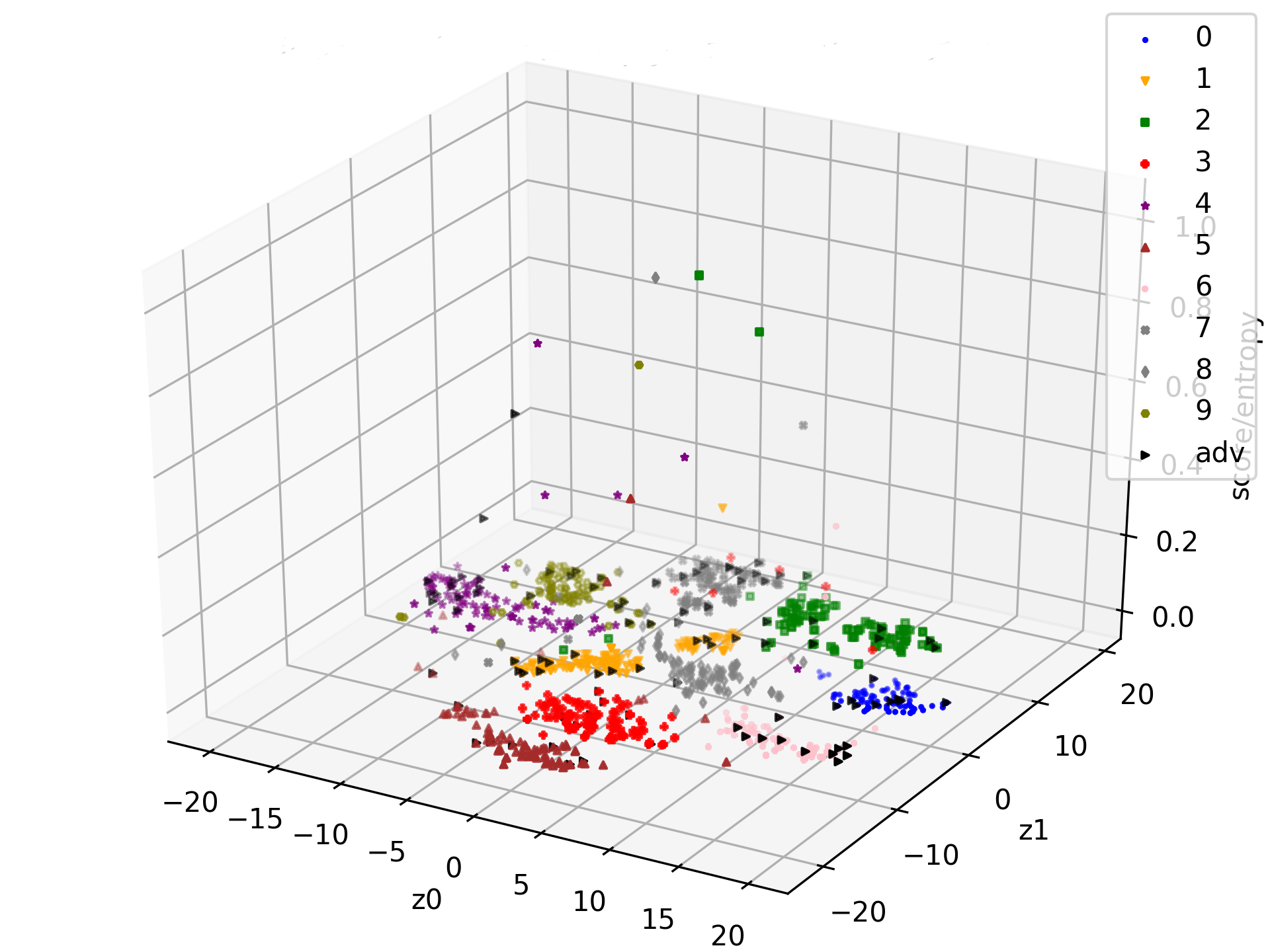}~~\includegraphics[width=0.45\textwidth]{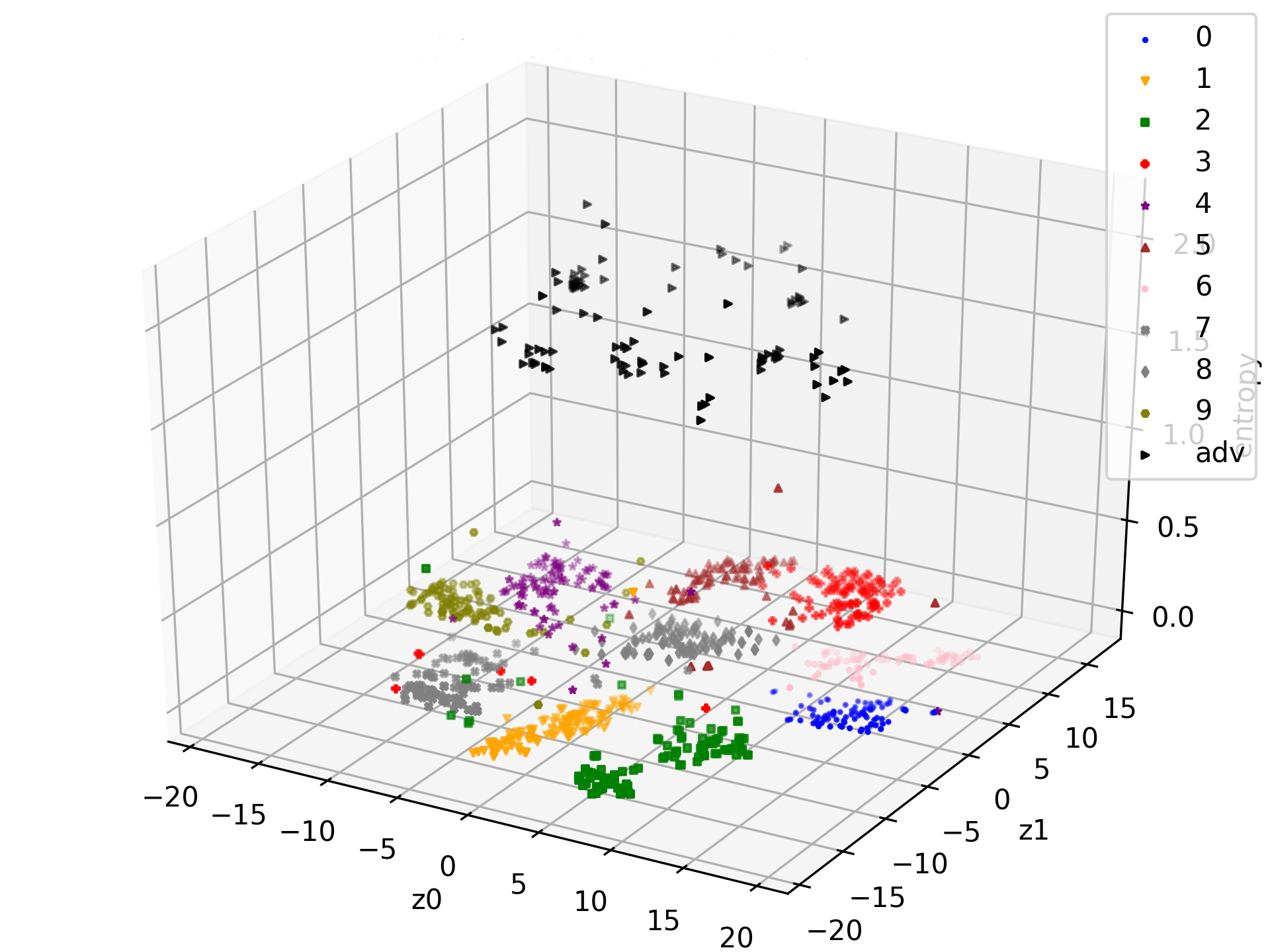}
\par\end{centering}
\caption{T-SNE visualization with entropy of prediction with entropy of prediction
probability. Black triangles are (negative) adversarial examples while
others are clean images. Left: ADR-None. Right: ADR+LC/GB\label{fig:neg-tsne-3d}}
\end{figure*}

\end{document}